\PassOptionsToPackage{expansion=false}{microtype}
\documentclass[]{jingdong}

\usepackage[utf8]{inputenc}
\usepackage[T1]{fontenc}

\usepackage{amsfonts}
\usepackage{amsmath}
\usepackage{amssymb}
\usepackage{dsfont}
\usepackage{algorithm}
\usepackage{algpseudocode}
\usepackage{booktabs}
\usepackage{colortbl}
\usepackage{tikz}
\usepackage{xcolor}
\usepackage{enumitem}
\usepackage{inconsolata}
\usepackage{listings}
\usepackage{nicefrac}
\usepackage{siunitx}
\usepackage{url}
\usepackage{xspace}
\usepackage{wrapfig}

\crefformat{section}{\S#2#1#3}
\Crefformat{section}{\S#2#1#3}
\crefmultiformat{section}{\S#2#1#3}{ and \S#2#1#3}{, \S#2#1#3}{ and \S#2#1#3}
\Crefmultiformat{section}{\S#2#1#3}{ and \S#2#1#3}{, \S#2#1#3}{ and \S#2#1#3}
\crefrangeformat{section}{\S#3#1#4 to \S#5#2#6}
\Crefrangeformat{section}{\S#3#1#4 to \S#5#2#6}

\definecolor{templatekeyword}{HTML}{E1251B}
\definecolor{templatestring}{HTML}{0B7A75}
\definecolor{templatecomment}{HTML}{6B7280}
\definecolor{templatecodebg}{HTML}{F7F7F8}
\definecolor{tableink}{HTML}{292824}
\definecolor{tablemuted}{HTML}{78756C}
\definecolor{tableclay}{HTML}{B7483D}
\definecolor{tablepaper}{HTML}{F0EEE6}
\definecolor{tablefocus}{HTML}{F5EBE4}
\definecolor{tableline}{HTML}{D9D5CB}

\makeatletter
\newcommand{\tableFocusRow}{%
    \rowcolor{tablefocus}%
    \begingroup
    \setbox\z@\hbox{0123456789}%
    \rule[-\dp\@arstrutbox]{0pt}{%
        \dimexpr 2\dp\@arstrutbox+\ht\z@-\dp\z@\relax}%
    \endgroup}
\makeatother

    \newcommand{\tableLowerIndicator}{%
        \tikz[baseline=2.2pt,x=1pt,y=1pt]{%
            \draw[tableink,line width=0.65pt,line cap=round]
                (1.4,9.6) .. controls (-0.3,7.5) and (-0.3,2.1) .. (1.4,0);
            \draw[tableink,line width=0.65pt,line cap=round]
                (10.6,9.6) .. controls (12.3,7.5) and (12.3,2.1) .. (10.6,0);
            \draw[tableink,line width=0.7pt,line cap=round,line join=round] (6,8.8)--(6,0.8);
            \draw[tableink,line width=0.7pt,line cap=round,line join=round] (3.8,3)--(6,0.8)--(8.2,3);}}

\lstdefinestyle{templatecode}{
  basicstyle=\ttfamily\small,
  columns=fullflexible,
  backgroundcolor=\color{templatecodebg},
  frame=single,
  rulecolor=\color{black!15},
  numberstyle=\tiny\color{gray},
  keywordstyle=\color{templatekeyword},
  commentstyle=\color{templatecomment},
  stringstyle=\color{templatestring},
  showstringspaces=false,
  tabsize=2,
  breaklines=true,
  breakatwhitespace=true,
  captionpos=b,
  xleftmargin=3.4pt,
  xrightmargin=3.4pt
}

\usepackage{needspace}

\renewcommand{\titlefont}{\sffamily\fontsize{12.7}{19}\selectfont}
\renewcommand{\authorfont}{\fontsize{10}{11}\selectfont}
\hypersetup{pdftitle={VideoJev: System One Model for Video Understanding and Interaction}}
\title{\texorpdfstring{%
  \raisebox{-4.5bp}{\includegraphics[width=\linewidth]{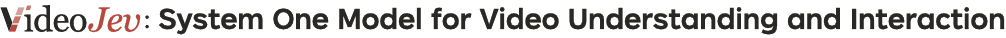}}%
}{VideoJev: System One Model for Video Understanding and Interaction}}
\author[*,1]{Yuhan Zhu}
\author[*,1]{Xiangyu Zeng}
\author[*,\ddagger,2]{Qingyi Si}
\author[*,3]{Chenxu Yang}
\author[*,4]{Yifei Li}
\author[5]{Luozheng Qin}
\let\equalauthorlist\authorlist
\renewcommand\authorlist{}
\author[3]{Junhao Zhou}
\author[3]{Dingyu Yao}
\author[2]{Jun Zhang}
\author[3]{Chuanyu Qin}
\author[6]{Zhiqiu Zhang}
\author[1]{Yuandong Yang}
\author[2]{Yuqi Xu}
\author[7]{Tianyu Zhang}
\author[8]{Hongcheng Guo}
\author[2]{Jiaqi Wang}
\author[2]{Nan Duan}
\author[\ddagger,\dagger,1]{Limin Wang}
\let\otherauthorlist\authorlist
\renewcommand\authorlist{\equalauthorlist\par\otherauthorlist}

\affiliation[1]{NJU}
\affiliation[2]{JD.com}
\affiliation[3]{UCAS}
\affiliation[4]{THU}
\affiliation[5]{XMU}
\affiliation[6]{SJTU}
\affiliation[7]{USTC}
\affiliation[8]{FDU}

\contribution[*]{Equal contribution}
\contribution[\ddagger]{Project lead}
\contribution[\dagger]{Corresponding author}
\newcommand{\projecthomepage}{https://mcg-nju.github.io/VideoJev}
\newcommand{\projectgithub}{https://github.com/MCG-NJU/VideoJev}
\newcommand{\projecthuggingface}{https://huggingface.co/collections/MCG-NJU/videojev}

\definecolor{projectlinkcolor}{HTML}{B7483D}
\definecolor{projectlinkink}{HTML}{292824}
\definecolor{projectlinkrule}{HTML}{DED6CB}
\definecolor{projectlinkicon}{HTML}{776B60}
\newcommand{\projectglyph}[1]{%
  \begin{tikzpicture}[x=1pt,y=1pt,baseline=-2pt,
      line width=.7pt,line cap=round,line join=round]
    \path[use as bounding box] (0,0) rectangle (12,11);
    \ifstrequal{#1}{project}{%
      \draw[projectlinkicon] (1,5.2) -- (6,9.7) -- (11,5.2)
        (2.5,4.9) -- (2.5,1) -- (9.5,1) -- (9.5,4.9);
      \draw[projectlinkcolor] (5,1) -- (5,4.1) -- (7,4.1) -- (7,1);
    }{}%
    \ifstrequal{#1}{code}{%
      \draw[projectlinkicon] (3.6,8.4) -- (.8,5.5) -- (3.6,2.6)
        (8.4,8.4) -- (11.2,5.5) -- (8.4,2.6);
      \draw[projectlinkcolor] (7.1,9) -- (4.9,2);
    }{}%
    \ifstrequal{#1}{models}{%
      \draw[projectlinkicon] (1.2,6) -- (6,3.7) -- (10.8,6)
        (1.2,3.6) -- (6,1.3) -- (10.8,3.6);
      \draw[projectlinkcolor] (1.2,8.4) -- (6,10.7) -- (10.8,8.4)
        -- (6,6.1) -- cycle;
    }{}%
  \end{tikzpicture}%
}

\newcommand{\projectentry}[3]{%
  \href{#1}{%
    \begin{tikzpicture}[baseline=(label.base)]
      \node[minimum width=74pt,minimum height=12pt,inner sep=0pt] (entry) {};
      \node[anchor=west,inner sep=0pt] at (entry.west)
        {\projectglyph{#2}};
      \node[anchor=west,inner sep=0pt,text=projectlinkink] (label)
        at ([xshift=18pt]entry.west) {\textls[65]{\MakeUppercase{#3}}};
      \draw[projectlinkcolor,line width=.6pt,line cap=round,line join=round]
        ([xshift=-5pt,yshift=-1.5pt]entry.east) --
        ([yshift=3pt]entry.east)
        ([xshift=-3.5pt,yshift=3pt]entry.east) --
        ([yshift=3pt]entry.east) --
        ([yshift=-.5pt]entry.east);
      \draw[projectlinkrule,line width=.4pt]
        ([yshift=-4.5pt]label.base -| entry.west) --
        ([yshift=-4.5pt]label.base -| entry.east);
      \draw[projectlinkcolor,line width=.65pt]
        ([yshift=-4.5pt]label.base -| entry.west) -- ++(8pt,0);
    \end{tikzpicture}%
  }%
}

\renewcommand{\checkdatalist}{%
  \begingroup
  \fontfamily{phv}\fontseries{m}\fontsize{7.5}{10}\selectfont
  \noindent\makebox[\linewidth][c]{%
    \scalebox{0.9}{\projectentry{\projecthomepage}{project}{Project}}%
    \hspace{24pt}%
    \scalebox{0.9}{\projectentry{\projectgithub}{code}{Code}}%
    \hspace{24pt}%
    \scalebox{0.9}{\projectentry{\projecthuggingface}{models}{Models}}%
  }%
  \endgroup
}

\abstract{Visual agents must turn video observations into timely decisions to track progress and choose what to do next. Building on the emerging System One paradigm, we introduce \textbf{VideoJev}, a family of models that cast video understanding and interaction as structured decision tasks. We adapt pretrained multimodal large language models (MLLMs) by replacing autoregressive answer decoding with a shared head that scores temporal intervals and textual options against video evidence. These scores yield event spans or answer choices, unifying temporal grounding and multiple-choice video question answering. This combines the semantic flexibility of MLLMs with efficient decisions that agents can use directly.
As an interaction unfolds, new goals and intermediate results prompt further queries about the same visual history. VideoJev therefore reuses visual computation through \emph{Cached inference} for queries as they arrive and \emph{Joint inference} for multiple queries in a single forward pass.
Across three temporal grounding benchmarks, VideoJev-4B and VideoJev-35B-A3B improve mean mIoU over their backbones by {\bfseries 6.7} and {\bfseries 10.8} points, respectively, with measured inference speedups of {\bfseries\boldmath $2.8$--$3.1\times$}. On four video QA benchmarks, both achieve performance comparable to their backbones in Thinking mode, with measured inference speedups exceeding {\boldmath $100\times$}. These results offer a unified, reusable decision interface for composing structured visual judgments into complex, open-ended workflows.}

\begin{document}
\maketitle
% !TEX root = ../main.tex
% First-page vector teaser; redraw with figures/draw_video_jev_teaser.py.
\par\noindent
\begin{minipage}{\linewidth}
\setlength{\parskip}{0pt}
\centering
\includegraphics[width=\linewidth]{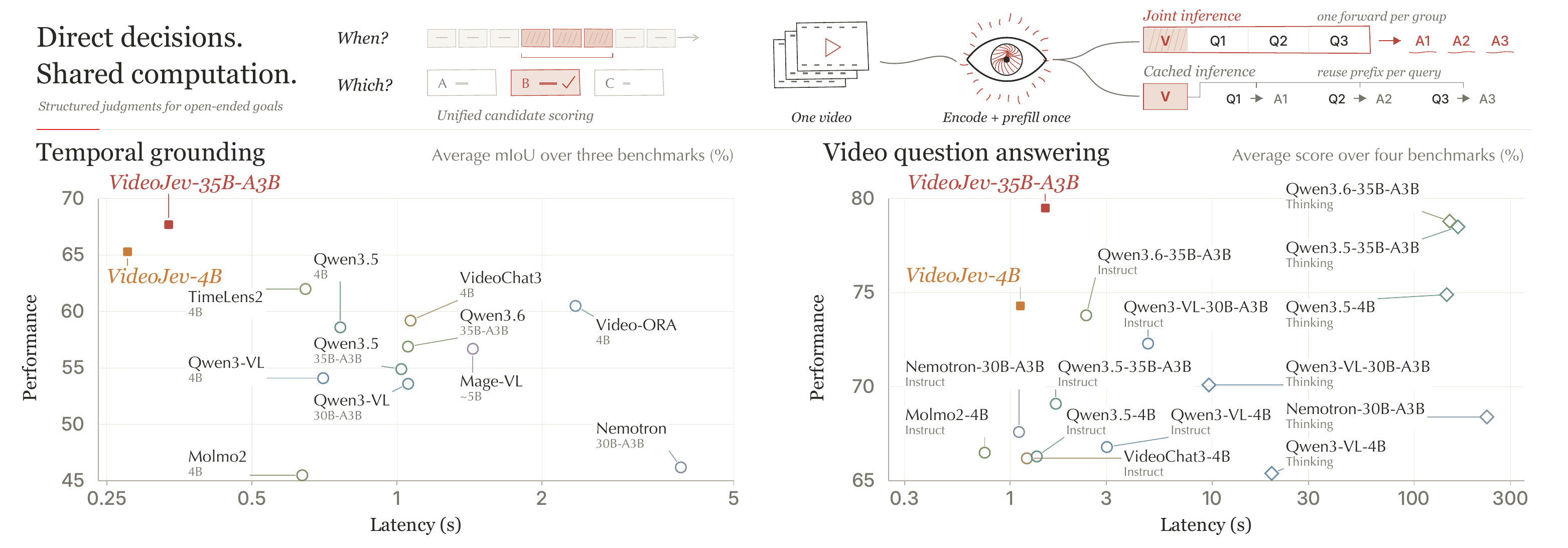}\par
\captionsetup{font=footnotesize,skip=2pt}
\captionof{figure}{\textbf{VideoJev: structured visual decisions with shared computation.} Top: temporal intervals and textual options share a candidate-scoring interface; \emph{Cached inference} reuses the video prefix for arriving queries, while \emph{Joint inference} scores each query group in one forward pass. Bottom: prediction quality versus model forward latency at both model scales, using the results in Tables~\ref{tab:temporal-grounding} and~\ref{tab:video-qa}. Left: mean grounding mIoU across three benchmarks and per-query latency on full Charades. Right: mean Video QA score across four benchmarks and per-question latency on full LVBench.}
\label{fig:teaser}
\end{minipage}
\par
\clearpage

% !TEX root = ../main.tex
\section{Introduction}

Visual intelligence enables systems to perceive, reason about, and interact with the world. For such systems, video understanding must support timely decisions that guide reasoning and action. An agent may need to locate evidence, check progress, or assess possible actions, often using the same visual history. For example, an assistant following a cooking procedure may check whether an ingredient has been added before choosing the next step. High-level goals can be decomposed into conditions to verify and options to evaluate, through predefined procedures or generated plans~\citep{ahn2022icanisay,yao2026benchmarkinglivingscreennativeguiagents}. This suggests a central premise: structured visual judgments can support open-ended behavior without requiring open-ended generation at every step.

Existing visual models offer complementary strengths. Discriminative vision models make efficient predictions but often depend on task-specific designs or fixed label spaces. Video multimodal large language models (MLLMs) provide broad semantic understanding and flexible language interfaces~\citep{li2025videochatflash,zeng2026videoo3}, yet typically express decisions through autoregressive generation. They generate timestamps for temporal grounding~\citep{ren2024timechattimesensitivemultimodallarge,huang2023vtimellmempowerllmgrasp,zhu2026timelens2generalistvideotemporal} or answer text for video question answering~\citep{li2024mvbenchcomprehensivemultimodalvideo,li2026videochat3fullyopenvideo}, even when the output is a compact, structured decision. Our goal is to combine efficient direct prediction with the semantic flexibility of pretrained MLLMs through a unified, language-conditioned interface.

The System One paradigm offers a foundation for this goal. Jev produces typed probabilistic judgments that software can use directly, without generating answers token by token~\citep{typesafe2026jev,typesafe2026systemone}. Recent work extends this idea to images~\citep{yu2026visualjevaccurateefficient,zhou2026textdecisionspixelsstudy}. Video introduces an additional requirement: decisions must reflect temporal structure while sharing expensive visual context across queries. When successive questions share the same observations, the cost of each additional judgment becomes central to responsive interaction.

We introduce \textbf{VideoJev}, a family of video System One models built around two principles: direct candidate scoring and reusable visual computation (Figure~\ref{fig:teaser}). Temporal grounding scores intervals for membership in a queried event and assembles them into a contiguous span. Multiple-choice video question answering (Video QA) scores textual options and selects the answer best supported by the video. Both tasks share a scoring architecture, while task-specific supervision and output rules preserve their different structures. This design draws on pretrained multimodal representations without autoregressive answer decoding.

To support repeated judgments, VideoJev organizes training and inference around a shared video prefix. In \emph{Cached inference}, the video is encoded and prefilled once, and queries reuse the cached context as they arrive. In \emph{Joint inference}, multiple queries follow the same prefix and receive separate scores in one causal forward pass. Training uses the same layout to share computation across questions.

We evaluate VideoJev-4B, a dense model, and VideoJev-35B-A3B, a mixture-of-experts (MoE) model, on three temporal grounding and four Video QA benchmarks. Relative to their Qwen3.5-4B and Qwen3.6-35B-A3B backbones in Instruct mode, the two models improve mean grounding mIoU by {\bfseries 6.7} and {\bfseries 10.8} points and mean QA scores by {\bfseries 8.0} and {\bfseries 5.7} points, respectively (Tables~\ref{tab:temporal-grounding} and~\ref{tab:video-qa}). Measured inference speedups reach {\bfseries\boldmath $2.8$--$3.1\times$} for grounding; on QA, both remain competitive with their Thinking counterparts while achieving over {\boldmath $100\times$} speedups. The timed generative baselines also reuse the video prefix.

With identical training data and settings, VideoJev-4B also exceeds autoregressive fine-tuning by 1.3 points in mean grounding mIoU and 1.9 points in mean QA score (Table~\ref{tab:same-data-control}), supporting the benefit of direct decision learning beyond data alone. These findings support learning structured judgments for visual agents.

Our contributions are threefold:
\begin{itemize}[leftmargin=*,nosep]
    \item We introduce VideoJev, a family of video System One models that unifies temporal grounding and multiple-choice Video QA through direct candidate scoring.
    \item We reuse visual computation across training queries and through two inference modes: \emph{Cached inference} for arriving queries and \emph{Joint inference} for queries processed together.
    \item We evaluate dense 4B and MoE 35B-A3B models across seven benchmarks, demonstrating strong prediction quality and inference efficiency, with a controlled comparison against autoregressive fine-tuning.
\end{itemize}

% !TEX root = ../main.tex
% Keep the section, subsection, and opening text on the same page.
\Needspace{9\baselineskip}
\section{VideoJev}

\subsection{A Unified Video Decision Framework}

\begin{figure}[t]
    \centering
    \includegraphics[width=\linewidth]{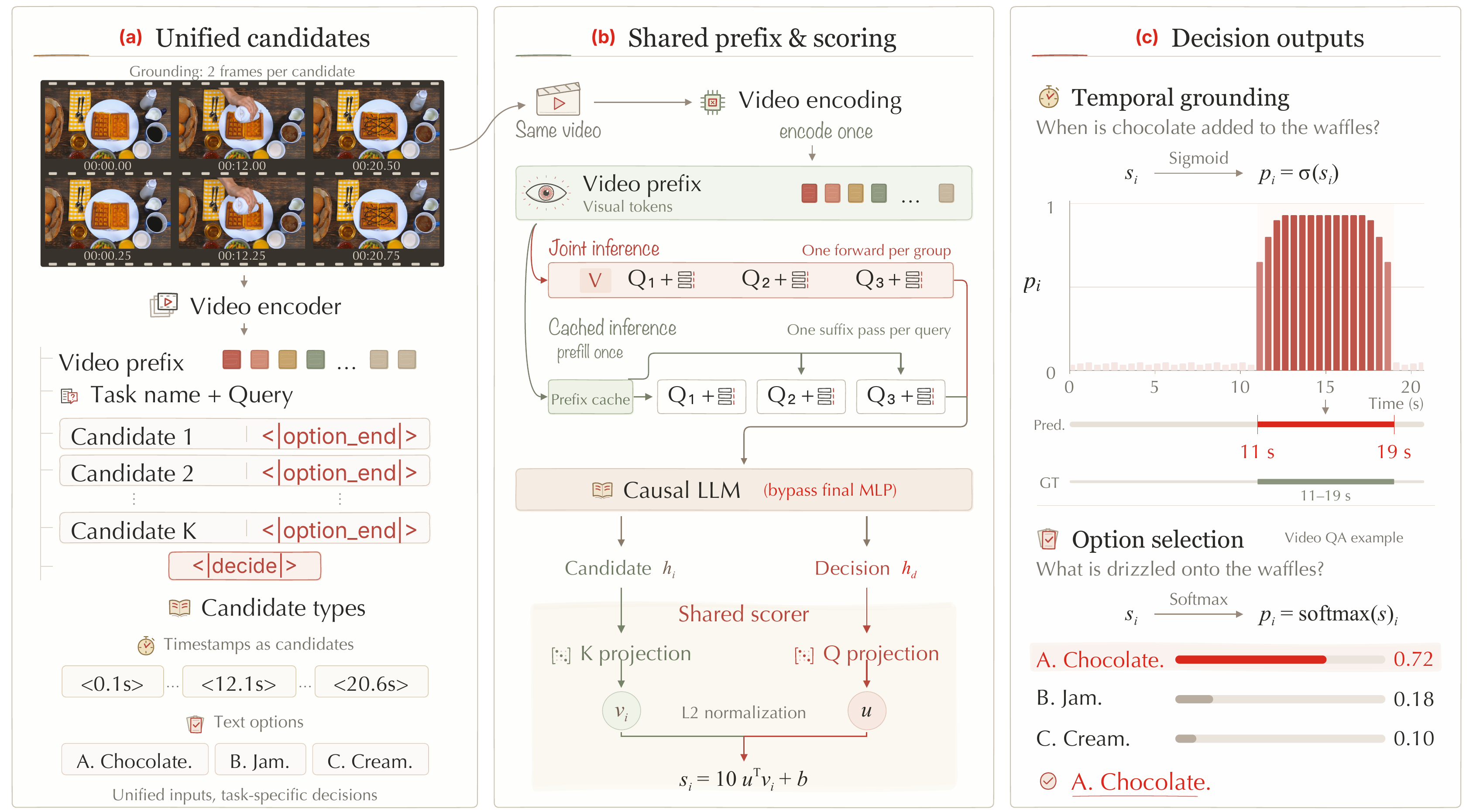}
    \caption{\textbf{VideoJev: direct candidate scoring with shared visual computation.} (a) Temporal intervals, represented by timestamps, and textual options share an input format. (b) \emph{Cached inference} prefills the video once and processes each query suffix separately; \emph{Joint inference} processes each query group in one causal language-model forward pass. Both modes share the backbone and scorer. (c) Temporal scores yield an event span; option scores select an answer. Scores and the reference interval are illustrative.}
    \label{fig:video-jev-overview}
\end{figure}

VideoJev scores temporal intervals and textual options with a shared head over pretrained multimodal representations. Figure~\ref{fig:video-jev-overview} summarizes the architecture, video-prefix reuse, and task-specific output rules.

\paragraph{Architecture.}
We adapt a pretrained MLLM with a causal language backbone and native video inputs. Following the observation that pre-MLP features support semantic similarity~\citep{zhu2025freeret}, we remove the final MLP and replace the language-model output head with a candidate scorer. Two new tokens, \texttt{<|option\_end|>} and \texttt{<|decide|>}, provide candidate readouts and a query-level decision readout. Both are initialized from the pretrained \texttt{<|im\_end|>} embedding. Separate linear projections map their hidden states to 512 dimensions and normalize them to unit length:
\begin{equation}
    u = \frac{W_qh_d+b_q}{\|W_qh_d+b_q\|_2}, \qquad
    v_i = \frac{W_kh_i+b_k}{\|W_kh_i+b_k\|_2}, \qquad
    s_i = 10\,u^\top v_i+b.
    \label{eq:cosine-decision}
\end{equation}
Here, $h_d$ and $h_i$ are the decision and candidate states; $W_q,W_k,b_q,b_k$ are learned projection parameters, and $\|\cdot\|_2$ is the Euclidean norm. Their normalized similarity gives logit $s_i$, with fixed scale 10 and learned scalar bias $b$.

\paragraph{Unifying temporal and option decisions.}
Given a video $V$, a language query $q$, and $K$ candidates $C=\{c_1,\ldots,c_K\}$, we place video tokens first, followed by the task instruction, query, and candidate list. The instruction specifies temporal grounding or multiple-choice Video QA. Each candidate ends with \texttt{<|option\_end|>}, and the list ends with \texttt{<|decide|>}. Causal modeling lets candidate readouts use the video, query, and preceding candidates; the decision readout sees the full list. Temporal candidates are timestamp texts copied from the video prefix into this list, each identifying a frame-pair interval. Their sigmoid scores indicate event membership, from which we decode a contiguous span. Option candidates are answer texts; softmax normalizes their logits into probabilities for selecting the highest-scoring option. Thus, candidates and output rules specify the task.

\paragraph{Reusable video computation.}
Causal modeling makes the video prefix independent of subsequent queries, allowing its computation to be shared. In \emph{Cached inference}, we encode and prefill the video once. Each arriving query branches from the cached prefix and requires one language-model forward over its query, candidates, and readout tokens. In \emph{Joint inference}, multiple queries and their candidate lists follow a single video prefix, with a separate \texttt{<|decide|>} readout for each query. One causal language-model forward pass yields all query scores. Later queries can access preceding query text and candidates, but no reference or predicted answers are appended. Training uses the same concatenated query layout to share video computation across questions for efficiency.
\subsection{Temporal Grounding through Interval Decisions}

An event may span many candidate intervals. Grounding therefore requires both identifying relevant intervals and covering the event as a whole. We train these complementary judgments with local membership and coverage supervision, then decode a single continuous span.

\paragraph{Learning temporal decisions.}
The backbone's twofold temporal compression groups frames into pairs. We partition the timeline into $K$ candidate intervals, one per pair. For interval $B_i=[\ell_i,r_i]$, $t_i$ is the mean of its two frame timestamps, and $\ell_i,r_i$ are its boundaries. Each internal boundary lies halfway between the previous pair's last frame and the next pair's first frame; $\ell_1=0$ and $r_K$ equals the video duration. Timestamp strings identify intervals; labels and boundaries use unrounded times. Let $G$ be the union of annotated event intervals. Candidate $i$ has label $y_i=\mathds{1}\{t_i\in G\}$ and membership probability $p_i=\sigma(s_i)$, where $\mathds{1}$ is the indicator function and $\sigma$ is the sigmoid function. Several candidates may belong to the same event, so we use multi-label binary cross-entropy to supervise each one independently:
\begin{equation}
    \mathcal{L}_{\mathrm{BCE}} = -\frac{1}{K}\sum_{i=1}^{K}
    \left[y_i\log p_i+(1-y_i)\log(1-p_i)\right].
\end{equation}
BCE supervises local membership but does not directly reward event coverage. We therefore add a soft intersection-over-union (IoU) loss over the positive intervals. With interval duration $d_i=r_i-\ell_i$, we define
\begin{equation}
    \mathcal{L}_{\mathrm{IoU}} = 1-
    \frac{\sum_i d_i p_i y_i}
    {\sum_i d_i\left[y_i+p_i(1-y_i)\right]}.
\end{equation}
The numerator rewards coverage of the event, while the denominator penalizes responses outside it. Weighting by duration accounts for unequal interval lengths.

Soft IoU is omitted when no midpoint is positive; BCE uses only valid candidates.

\paragraph{Decoding a temporal interval.}
To combine interval scores into an event span, we adjust logits by a probability threshold $\tau$ and select the nonempty contiguous range with the largest sum. We fix $\tau=0.25$ across both model sizes and all inference modes in every experiment, with smoothing window 1:
\begin{equation}
    (a^\star,b^\star)=\arg\max_{1\le a\le b\le K}
    \sum_{i=a}^{b}\left[s_i-\log\frac{\tau}{1-\tau}\right].
\end{equation}
Here, $a$ and $b$ index the first and last candidates in a range, and $a^\star$ and $b^\star$ identify the selected range. Kadane's algorithm finds this range in $O(K)$ time. The selected candidates' interval boundaries give the continuous prediction $[\ell_{a^\star},r_{b^\star}]$. If all adjusted logits are negative, the decoder returns the interval of the strongest candidate. We then apply lightweight boundary postprocessing to obtain a valid, nonempty interval.

\subsection{Video QA through Option Selection}

Multiple-choice Video QA requires selecting one textual option rather than identifying multiple positive intervals.

\paragraph{Learning candidate preferences.}
For a question with $K$ answer options, the shared head assigns logit $s_i$ to option $i$. Let $y$ be the index of the correct option. Since exactly one answer is correct, we normalize the logits with softmax and train with cross-entropy:
\begin{equation}
    p_i = \frac{\exp(s_i)}{\sum_{j=1}^{K}\exp(s_j)}, \qquad
    \mathcal{L}_{\mathrm{CE}} = -\log p_y,
\end{equation}
where $p_i$ is the probability of option $i$.

\paragraph{Reducing position bias.}
To discourage shortcuts based on option position, we apply a reproducible, answer-independent permutation to each eligible training question and update its correct-answer index. Questions referring to option positions or earlier turns retain their original order to preserve meaning. At inference, we keep the supplied order and select the option with the highest $p_i$.

\subsection{Training Recipe}

We instantiate VideoJev with Qwen3.5-4B and Qwen3.6-35B-A3B. Scoring-head projection weights are initialized from a Gaussian with standard deviation 0.005; biases start at zero.

\paragraph{Training data.}
We train temporal membership on TimeLens-100K~\citep{Zhang_2026_CVPR} and TimeLens2-93K~\citep{zhu2026timelens2generalistvideotemporal}, converting annotated intervals into candidate-level labels. Option selection uses an internally curated corpus of approximately 240K multiple-choice questions over 19K videos. The content includes gaming, sports, cooking, music and dance, outdoor activities, and crafts. Questions cover entity and action recognition, spatiotemporal reasoning, information retrieval, event understanding, counting, OCR, and summarization. Videos range from 4 minutes to nearly 2 hours, and each question has four or five answer options. We group queries from the same video around a shared prefix.

\paragraph{Two-stage training.}
Inspired by projector alignment in MLLMs~\citep{liu2023visualinstruction}, we first train the new scorer against fixed pretrained features, then adapt the language backbone. This gives the scorer a stable representation space during its initial updates. Stage one trains only the head on 30\% of the videos, with the backbone and new marker embeddings frozen. The grounding and QA objectives are
\begin{equation}
    \mathcal{L}_{\mathrm{TG}} = \mathcal{L}_{\mathrm{BCE}}, \qquad
    \mathcal{L}_{\mathrm{QA}} = \mathcal{L}_{\mathrm{CE}}.
\end{equation}
Stage two trains the language backbone, marker embeddings, and aligned head on the remaining 70\% of the videos. Soft IoU adds event-level coverage supervision to grounding, while QA retains cross-entropy:
\begin{align}
    \mathcal{L}_{\mathrm{TG}} &= \mathcal{L}_{\mathrm{BCE}}
    +2\,\mathcal{L}_{\mathrm{IoU}}, \label{eq:joint-grounding-loss} \\
    \mathcal{L}_{\mathrm{QA}} &= \mathcal{L}_{\mathrm{CE}}. \label{eq:joint-qa-loss}
\end{align}
Task losses are averaged equally over questions. The visual encoder and merger remain frozen throughout. Each stage runs for one epoch with a separate optimizer and learning-rate schedule.

% !TEX root = ../main.tex
\section{Experiments}

We evaluate prediction quality and inference efficiency on grounding and multiple-choice Video QA, then examine model design and confidence. Unless otherwise stated, VideoJev uses \emph{Joint inference}.

\subsection{Experimental Setup}

\paragraph{Training configuration.}
We train both model sizes on 64 NVIDIA H200 GPUs with BF16 computation and a global batch size of 64 packed sequences. Training videos are sampled at 2 fps, with a maximum sequence length of 144K tokens and an 80\% visual-token budget. We use AdamW with weight decay 0.01 and gradient clipping at 1.0. Each stage uses 5\% warm-up followed by cosine decay to 10\% of its peak learning rate. The head learning rate is $3\times10^{-4}$ during head alignment and $10^{-4}$ during joint adaptation. In the latter stage, the language-backbone learning rate is $5\times10^{-6}$ for 4B and $2\times10^{-6}$ for 35B-A3B. The two stages run for 160 and 375 optimizer updates, respectively. We use DeepSpeed ZeRO-2 for 4B and ZeRO-3 for 35B-A3B.

\paragraph{Temporal grounding benchmarks and metrics.}
We evaluate on the TimeLens-Bench refined annotations~\citep{Zhang_2026_CVPR} of Charades-STA~\citep{Gao_2017_ICCV}, ActivityNet Captions~\citep{Krishna_2017_ICCV}, and QVHighlights~\citep{lei2021neurips-detecting}. For each query, we report rank-one recall at temporal IoU thresholds 0.3, 0.5, and 0.7 (R1@0.3/0.5/0.7), together with mean temporal IoU (mIoU). VideoJev uses a fixed decoding threshold $\tau=0.25$ and smoothing window 1 across both model sizes and all inference modes.

\paragraph{Video QA benchmarks and metrics.}
We evaluate multiple-choice Video QA on the LVBench~\citep{wang2025lvbenchextremelongvideo} test set (1,549 questions), Video-MME~\citep{fu2025videommefirstevercomprehensiveevaluation} test set (2,700 questions), LongVideoBench~\citep{wu2024longvideobenchbenchmarklongcontextinterleaved} validation set (1,337 questions), and the seven multiple-choice tasks of MLVU~\citep{zhou2025mlvubenchmarkingmultitasklong} development (2,174 questions). Inputs contain video frames only, without subtitles or audio. We report accuracy on LVBench, Video-MME, and LongVideoBench, and M-Avg on MLVU.

\paragraph{Inference latency.}
We measure grounding latency on the complete Charades test set (1,313 videos; 3,363 queries) and QA latency on the complete LVBench test set (103 videos; 1,549 questions), using H200 GPUs at batch size 1. Both tasks measure model forward time only, excluding preprocessing and data transfers. VideoJev uses \emph{Joint inference}, encoding each video once and scoring its queries together. Generative baselines share visual encoding and prefix prefill, then generate answers from independent query branches. We divide accumulated forward time, including visual encoding, prefill, and all autoregressive decoding steps, by the number of queries. Generations reaching the output limit are included. Grounding uses 4 fps inputs and at most $640\times640$ pixels per frame; QA models retain their native visual settings. Appendix~\ref{app:latency} gives the timing details.

\begin{table}[!t]
    \begingroup
    \color{tableink}
    \colorlet{tablefocus}{tableclay!5}
    \hypersetup{citecolor=tableclay,urlcolor=tableclay}
    % Table 1 palette and small editorial accents, inspired by Figure 1.
    \definecolor{vtgsand}{HTML}{A8844C}
    \definecolor{vtgsage}{HTML}{778269}
    \definecolor{vtgmoe}{HTML}{6D7F89}
    \definecolor{vtgsandwash}{HTML}{FAF6ED}
    \definecolor{vtgsagewash}{HTML}{F3F5EF}
    \definecolor{vtgclaywash}{HTML}{FCF3EF}
    \definecolor{vtgsummary}{HTML}{F2ECE2}
    \definecolor{vtgpaper}{HTML}{FCFBF8}
    \newcommand{\vtgPlainHeading}[3]{%
        \begin{tikzpicture}[baseline=(label.base)]
        \node[rounded corners=3.5pt,text=tableink,inner xsep=12pt,inner ysep=5pt,font=\bfseries] (label) {#1};
        \draw[#2,line cap=round,line width=0.85pt]
            ([xshift=9pt,yshift=1.4pt]label.south west)
            .. controls ([xshift=25pt,yshift=2.1pt]label.south west)
            and ([xshift=-25pt,yshift=0.7pt]label.south east)
            .. ([xshift=-9pt,yshift=1.5pt]label.south east);
        \end{tikzpicture}}
    \newcommand{\vtgHeading}[3]{\vtgPlainHeading{#1\textsuperscript{TL}}{#2}{#3}}
    \newcommand{\vtgTag}[2]{%
        \tikz[baseline=(label.base)]\node[rounded corners=3pt,fill=#2!14,text=#2,inner xsep=5pt,inner ysep=2pt,font=\bfseries] (label) {#1};}
    \newcommand{\tblSecond}[1]{\uline{#1}}
    \centering
    \caption{\textbf{Video temporal grounding.} Generative models use Instruct mode by default; VideoJev uses Joint inference. The superscript \textsuperscript{TL} marks TimeLens-Bench refined annotations. \textbf{Bold} marks the best value and \uline{underline} the second best. Appendix~\ref{app:protocol} details the evaluation settings.}
    \label{tab:temporal-grounding}
    \setlength{\tabcolsep}{3.6pt}
    \renewcommand{\arraystretch}{1.22}
    \resizebox{\textwidth}{!}{%
    \fontfamily{jdlangzhengtiregular}\selectfont
    \begin{tabular}{@{}lc!{\hspace{4pt}\color{tableline}\vrule width 0.45pt\hspace{4pt}}*{3}{c}c*{3}{c}c*{3}{c}c!{\hspace{4pt}\color{tableline}\vrule width 0.45pt\hspace{4pt}}c@{}}
        \arrayrulecolor{jingdongred}\toprule[0.85pt]
        \arrayrulecolor{tableline}
        \rowcolor{white}
        & \vtgPlainHeading{Latency \tableLowerIndicator}{vtgmoe}{vtgsagewash} & \multicolumn{4}{c}{\vtgHeading{Charades}{vtgsand}{vtgsandwash}} & \multicolumn{4}{c}{\vtgHeading{ActivityNet}{vtgsage}{vtgsagewash}} & \multicolumn{4}{c}{\vtgHeading{QVHighlights}{tableclay}{vtgclaywash}} & \\
        \addlinespace[1.5pt]
        \rowcolor{white}
        \cellcolor{white}\multirow{-2}{*}[3pt]{{\bfseries Model}} & \cellcolor{white}seconds & R1@.3 & R1@.5 & R1@.7 & mIoU & R1@.3 & R1@.5 & R1@.7 & mIoU & R1@.3 & R1@.5 & R1@.7 & mIoU & {\bfseries Avg.} \\
        \midrule[0.55pt]
        \rowcolor{white}\multicolumn{15}{@{}l}{{\color{tablemuted}\bfseries Proprietary models}} \\
        GPT-5~\citep{OAI:gpt-5-system-card} & -- & 59.3 & 42.0 & 22.0 & 40.5 & 57.4 & 44.9 & 30.4 & 42.9 & 72.4 & 60.4 & 46.4 & 56.8 & 46.7 \\
        Gemini-2.5-Pro~\citep{Gemini2025Comanici} & -- & 74.1 & 61.1 & 34.0 & 52.8 & 72.3 & 64.2 & 47.1 & 58.1 & 84.1 & 75.9 & 61.1 & 70.4 & 60.4 \\
        \midrule[0.45pt]
        \rowcolor{white}\multicolumn{15}{@{}l}{{\color{tablemuted}\bfseries Open-source dense models (4B language backbone)}} \\
        % InternVL3.5-4B~\citep{wang2025internvl35advancingopensourcemultimodal} & -- & 24.6 & 14.1 & 6.2 & 16.0 & 20.0 & 12.7 & 8.2 & 14.9 & 25.5 & 15.8 & 6.2 & 17.7 & 16.2 \\
        Molmo2-4B~\citep{clark2026molmo2openweightsdata} & \tblSecond{0.636} & 44.4 & 31.1 & 18.5 & 34.7 & 50.8 & 40.3 & 29.9 & 40.9 & 73.7 & 62.6 & 51.8 & 60.8 & 45.5 \\
        Qwen3-VL-4B~\citep{bai2025qwen3vltechnicalreport} & 0.703 & 71.8 & 53.8 & 28.7 & 49.4 & 67.4 & 54.8 & 37.6 & 50.7 & 78.4 & 66.4 & 51.1 & 62.3 & 54.1 \\
        Mage-VL~\citep{yang2026magevl} & 1.437 & 75.5 & 63.1 & 36.1 & 55.0 & 65.1 & 57.0 & 39.7 & 51.6 & 76.4 & 68.1 & 53.6 & 63.4 & 56.7 \\
        Qwen3.5-4B~\citep{qwen35blog} & 0.763 & 75.2 & 56.6 & 31.5 & 52.3 & 72.6 & 61.6 & 42.6 & 55.6 & 82.9 & 73.1 & 58.1 & 67.9 & 58.6 \\
        VideoChat3-4B~\citep{li2026videochat3fullyopenvideo} & 1.067 & 78.4 & 64.9 & 35.9 & 56.1 & 70.1 & 60.5 & 42.2 & 54.6 & 81.0 & 72.5 & 56.8 & 67.0 & 59.2 \\
        Video-ORA-4B~\citep{li2026orarl} & 2.347 & 77.1 & 64.9 & \tblSecond{37.9} & 56.8 & 69.3 & 60.1 & 43.8 & 57.3 & 81.4 & 72.6 & 58.6 & 67.5 & 60.5 \\
        TimeLens2-4B~\citep{zhu2026timelens2generalistvideotemporal} & 0.646 & \tblSecond{80.3} & \tblSecond{66.6} & 37.5 & \tblSecond{57.7} & \tblSecond{74.6} & \tblSecond{65.5} & \tblSecond{47.2} & \tblSecond{59.0} & \tblSecond{83.3} & \tblSecond{75.2} & \tblSecond{60.0} & \tblSecond{69.3} & \tblSecond{62.0} \\
        \tableFocusRow{\color{tableclay}\bfseries VideoJev-4B} & {\bfseries 0.276} & {\bfseries 82.8} & {\bfseries 69.4} & {\bfseries 40.3} & {\bfseries 60.1} & {\bfseries 78.8} & {\bfseries 70.5} & {\bfseries 51.4} & {\bfseries 62.8} & {\bfseries 87.3} & {\bfseries 79.4} & {\bfseries 64.8} & {\bfseries 73.2} & {\bfseries 65.3} \\
        \midrule[0.45pt]
        \rowcolor{white}\multicolumn{15}{@{}l}{{\color{tablemuted}\bfseries Open-source MoE models (30-35B total, 3B active)}} \\
        Nemotron-3-Nano-Omni-30B-A3B~\citep{nvidia2026nemotronomni} & 3.882 & 65.2 & \tblSecond{49.9} & \tblSecond{26.6} & 45.3 & 54.7 & 43.5 & 27.9 & 40.6 & 66.7 & 55.7 & 41.2 & 52.6 & 46.2 \\
        Qwen3-VL-30B-A3B~\citep{bai2025qwen3vltechnicalreport} & 1.055 & 70.3 & 46.5 & 25.1 & \tblSecond{48.1} & 65.7 & 51.1 & 36.5 & 49.4 & 79.3 & 67.6 & 52.6 & 63.2 & 53.6 \\
        Qwen3.5-35B-A3B~\citep{qwen35blog} & \tblSecond{1.021} & 71.5 & 44.2 & 21.1 & 46.0 & 72.1 & 57.3 & 40.4 & 54.4 & 80.3 & 67.3 & 53.2 & 64.3 & 54.9 \\
        Qwen3.6-35B-A3B~\citep{qwen36_35b_a3b} & 1.054 & \tblSecond{72.6} & 47.0 & 23.2 & 47.3 & \tblSecond{73.9} & \tblSecond{59.8} & \tblSecond{42.4} & \tblSecond{56.0} & \tblSecond{83.3} & \tblSecond{72.5} & \tblSecond{57.6} & \tblSecond{67.5} & \tblSecond{56.9} \\
        \tableFocusRow{\color{tableclay}\bfseries VideoJev-35B-A3B} & {\bfseries 0.336} & {\bfseries 86.1} & {\bfseries 74.8} & {\bfseries 44.5} & {\bfseries 63.7} & {\bfseries 81.0} & {\bfseries 72.5} & {\bfseries 54.4} & {\bfseries 65.5} & {\bfseries 88.0} & {\bfseries 79.9} & {\bfseries 65.5} & {\bfseries 74.0} & {\bfseries 67.7} \\
        \arrayrulecolor{tableink}\bottomrule[0.85pt]
    \end{tabular}%
    }
    \endgroup
\end{table}

\subsection{Video Temporal Grounding}

Table~\ref{tab:temporal-grounding} shows consistent gains across the three grounding benchmarks. VideoJev-4B reaches 65.3 mean mIoU, improving on Qwen3.5-4B by {\bfseries 6.7} points and TimeLens2-4B by 3.3 points. Against TimeLens2-4B, it also raises R1@0.7 by 2.8--4.8 points across the three datasets. Gains under this stricter overlap criterion indicate more precise temporal localization, supporting interval membership scoring as an effective way to recover event spans.

VideoJev-35B-A3B raises mean mIoU to 67.7, {\bfseries 10.8} points above its Qwen3.6-35B-A3B backbone. Scaling from VideoJev-4B adds 2.4 points with only a 22\% increase in measured latency. Both sizes lead all reported grounding metrics and achieve the lowest measured latency within their respective open-source groups, with {\bfseries\boldmath $2.8$--$3.1\times$} speedups over their backbones. Since the generative baselines also share video prefill, these speedups highlight the value of avoiding autoregressive timestamp generation when repeated queries draw on the same visual context.

% Keep the two main comparison tables at the tops of separate pages.
\FloatBarrier
\suppressfloats[t]
\begin{table}[t]
    \begingroup
    \color{tableink}
    \colorlet{tablefocus}{tableclay!5}
    \hypersetup{citecolor=tableclay,urlcolor=tableclay}
    \centering
    \newcommand{\tblSecond}[1]{\uline{#1}}
    \caption{\textbf{Multiple-choice Video QA.} Molmo2 uses a 32K context; Nemotron uses 128K. \textbf{Bold} marks the best value and \uline{underline} the second best. Appendix~\ref{app:protocol} details the evaluation settings.}
    \label{tab:video-qa}
    \setlength{\tabcolsep}{4pt}
    \renewcommand{\arraystretch}{1.16}
    \resizebox{\textwidth}{!}{%
    \fontfamily{jdlangzhengtiregular}\selectfont
    \begin{tabular}{@{}lc!{\hspace{3pt}\color{tableline}\vrule width 0.45pt\hspace{3pt}}*{4}{c}!{\hspace{3pt}\color{tableline}\vrule width 0.45pt\hspace{3pt}}c@{}}
        \arrayrulecolor{jingdongred}\toprule[0.85pt]
        \arrayrulecolor{tableline}
        \rowcolor{white}
        & {\bfseries Latency \tableLowerIndicator} & {\bfseries LVBench} & {\bfseries Video-MME} & {\bfseries LongVideoBench} & {\bfseries MLVU} & {\bfseries Avg.} \\
        \rowcolor{white}
        \multirow{-2}{*}{{\bfseries Model}} & {\scriptsize seconds} & {\scriptsize test; w/o sub.} & {\scriptsize test; w/o sub.} & {\scriptsize val; w/o sub.} & {\scriptsize dev MCQ; M-Avg} & {\scriptsize 4-bench mean} \\
        \midrule[0.55pt]
        \rowcolor{white}\multicolumn{7}{@{}l}{{\color{tablemuted}\bfseries Proprietary models}} \\
        GPT-5.2~\citep{qwen35blog} & -- & 73.7 & 85.8 & -- & 85.6 & -- \\
        Gemini-3-Pro~\citep{qwen35blog} & -- & 76.2 & 87.7 & -- & 83.0 & -- \\
        Claude-Opus-4.5~\citep{qwen35blog} & -- & 57.3 & 81.4 & -- & 81.7 & -- \\
        Seed2.0-Pro~\citep{seed20modelcard} & -- & 76.4 & -- & 80.3 & -- & -- \\
        \midrule[0.45pt]
        \rowcolor{white}\multicolumn{7}{@{}l}{{\color{tablemuted}\bfseries Open-source dense models (4B language backbone)}} \\
        Molmo2-4B~\citep{clark2026molmo2openweightsdata} & {\bfseries 0.748} & 53.9 & 69.6 & \tblSecond{68.0} & 74.3 & 66.5 \\
        VideoChat3-4B~\citep{li2026videochat3fullyopenvideo} & 1.210 & 56.7 & 70.1 & 64.1 & 73.8 & 66.2 \\
        Qwen3-VL-4B\textsubscript{Instruct}~\citep{bai2025qwen3vltechnicalreport} & 3.012 & 56.2 & 69.3 & 66.5 & 75.3 & 66.8 \\
        Qwen3-VL-4B\textsubscript{Thinking}~\citep{bai2025qwen3vltechnicalreport} & 19.713 & 53.5 & 68.9 & 63.4 & 75.7 & 65.4 \\
        Qwen3.5-4B\textsubscript{Instruct}~\citep{qwen35blog} & 1.355 & 60.0 & 70.6 & 61.2 & 73.3 & 66.3 \\
        Qwen3.5-4B\textsubscript{Thinking}~\citep{qwen35blog} & 144.929 & \tblSecond{66.4} & {\bfseries 76.9} & {\bfseries 73.6} & {\bfseries 82.8} & {\bfseries 74.9} \\
        \tableFocusRow{\color{tableclay}\bfseries VideoJev-4B} & \tblSecond{1.123} & {\bfseries 67.9} & \tblSecond{75.0} & {\bfseries 73.6} & \tblSecond{80.6} & \tblSecond{74.3} \\
        \midrule[0.45pt]
        \rowcolor{white}\multicolumn{7}{@{}l}{{\color{tablemuted}\bfseries Open-source MoE models (30-35B total, 3B active)}} \\
        Nemotron-3-Nano-Omni-30B-A3B\textsubscript{Instruct}~\citep{nvidia2026nemotronomni} & {\bfseries 1.106} & 54.9 & 70.8 & 67.5 & 77.2 & 67.6 \\
        Nemotron-3-Nano-Omni-30B-A3B\textsubscript{Thinking}~\citep{nvidia2026nemotronomni} & 229.095 & 52.2 & 72.2 & 69.6 & 79.6 & 68.4 \\
        Qwen3-VL-30B-A3B\textsubscript{Instruct}~\citep{bai2025qwen3vltechnicalreport} & 4.817 & 62.5 & 74.5 & 70.9 & 81.3 & 72.3 \\
        Qwen3-VL-30B-A3B\textsubscript{Thinking}~\citep{bai2025qwen3vltechnicalreport} & 9.637 & 59.2 & 73.3 & 69.1 & 78.9 & 70.1 \\
        Qwen3.5-35B-A3B\textsubscript{Instruct}~\citep{qwen35blog} & 1.683 & 58.8 & 74.4 & 65.7 & 77.4 & 69.1 \\
        Qwen3.5-35B-A3B\textsubscript{Thinking}~\citep{qwen35blog} & 164.730 & \tblSecond{71.4} & {\bfseries 82.5} & 74.5 & 85.6 & 78.5 \\
        Qwen3.6-35B-A3B\textsubscript{Instruct}~\citep{qwen36_35b_a3b} & 2.379 & 66.0 & 77.1 & 72.4 & 79.5 & 73.8 \\
        Qwen3.6-35B-A3B\textsubscript{Thinking}~\citep{qwen36_35b_a3b} & 149.777 & \tblSecond{71.4} & {\bfseries 82.5} & \tblSecond{74.9} & {\bfseries 86.2} & \tblSecond{78.8} \\
        \tableFocusRow{\color{tableclay}\bfseries VideoJev-35B-A3B} & \tblSecond{1.493} & {\bfseries 73.4} & \tblSecond{81.8} & {\bfseries 77.1} & \tblSecond{85.9} & {\bfseries 79.5} \\
        \arrayrulecolor{tableink}\bottomrule[0.85pt]
    \end{tabular}%
    }
    \endgroup
\end{table}

\subsection{Option Selection on Video QA Benchmarks}

All VideoJev long-video QA evaluations in this paper use a fixed 256K (262,144-token) context budget across benchmarks, model sizes, and inference modes. We sample the full video at a requested 2 fps, uniformly subsampling when needed to cap the input at 2,048 frames. Frames contain at most $640\times640$ pixels; resolution is reduced as needed to keep visual tokens within 80\% of the context budget. Table~\ref{tab:video-qa} groups baselines by language-backbone scale and distinguishes Instruct and Thinking modes where reported.

In Table~\ref{tab:video-qa}, VideoJev-4B and VideoJev-35B-A3B improve four-benchmark mean scores over their Instruct backbones by {\bfseries 8.0} and {\bfseries 5.7} points, respectively. Both lead their open-source groups on LVBench, and the MoE model achieves the highest group mean. The gains extend to every QA benchmark.

Direct scoring also remains competitive with explicit reasoning. VideoJev-4B is within 0.6 points of Qwen3.5-4B Thinking in mean score, while VideoJev-35B-A3B exceeds Qwen3.6-35B-A3B Thinking by 0.7 points. Both outperform their Thinking counterparts on LVBench and match or improve on LongVideoBench; Thinking retains an edge on Video-MME and MLVU. Measured speedups over these Thinking baselines are {\boldmath $129\times$} and {\boldmath $100\times$}, respectively. Section~\ref{sec:analysis} further compares direct-option and reasoning baselines.

\subsection{Ablation Study}

We control for training data, then examine inference modes, task composition, supervision, and design choices. Architectural, objective, and schedule ablations use VideoJev-4B with the same complete test sets, \emph{Joint inference}, and grounding decoder. QA ablations and the autoregressive control retain the fixed 256K-token context; tables report changes from the marked baseline.

\begingroup
\definecolor{ablationsage}{HTML}{778269}
\colorlet{tablefocus}{tableclay!5}
\newcommand{\ablationHeading}[2]{%
    \begin{tikzpicture}[baseline=(label.base)]
    \node[text=tableink,inner xsep=12pt,inner ysep=5pt,font=\bfseries] (label) {#1};
    \draw[#2,line cap=round,line width=0.85pt]
        ([xshift=9pt,yshift=1.4pt]label.south west)
        .. controls ([xshift=25pt,yshift=2.1pt]label.south west)
        and ([xshift=-25pt,yshift=0.7pt]label.south east)
        .. ([xshift=-9pt,yshift=1.5pt]label.south east);
    \end{tikzpicture}}
% Baselines need only a score; a placeholder delta would lift it above the label.
% Center each actual score/delta block on the same optical axis as single-line text.
\newsavebox{\ablationReference}
\newsavebox{\ablationContent}
\newcommand{\ablationScore}[3]{%
    \ifstrequal{#2}{\textemdash}{#1}{%
        \sbox{\ablationReference}{0123456789}%
        \sbox{\ablationContent}{%
            \raisebox{\dimexpr(\ht\ablationReference-\dp\ablationReference-\height+\depth)/2\relax}{%
                \shortstack[c]{#1\\[1pt]{\scriptsize\color{#3}#2}}}}%
        \raisebox{0pt}[\dimexpr\ht\ablationContent+2pt\relax][\dimexpr\dp\ablationContent+2pt\relax]{%
            \usebox{\ablationContent}}}}

\paragraph{Isolating VideoJev's architectural gains.}
To test whether the gains arise solely from our training data, we fine-tune an ordinary Qwen3.5-4B autoregressive model on exactly the same post-training data as VideoJev. It retains the native final MLP and language-model head and learns to generate temporal intervals or answer letters. Both models use single-stage training on 100\% of the data; VideoJev results are copied from the single-stage variant in Table~\ref{tab:schedule-ablation}. Table~\ref{tab:same-data-control} shows that VideoJev improves mean grounding and QA scores by 1.3 and 1.9 points. The gains under matched data and schedules support an architectural benefit: directly supervising temporal membership and option preferences aligns the readout with the required decisions, enabling more effective use of the same supervision. Appendix~\ref{app:same-data-control} gives the key settings.

\begin{table}[!htb]
    \centering
    \caption{\textbf{Isolating architectural gains with an autoregressive control.} VideoJev-4B and an autoregressive Qwen3.5-4B are both fine-tuned in a single stage on exactly the same post-training data. \ensuremath{\dagger} denotes the VideoJev-4B Single-stage (100\%) variant from Table~\ref{tab:schedule-ablation}. The two models share the same training and evaluation settings.}
    \label{tab:same-data-control}
    \begingroup
    \color{tableink}
    \setlength{\tabcolsep}{3.6pt}
    \renewcommand{\arraystretch}{1.16}
    \resizebox{\linewidth}{!}{%
    \fontfamily{jdlangzhengtiregular}\selectfont
    \begin{tabular}{@{}l!{\hspace{3pt}\color{tableline}\vrule width 0.45pt\hspace{3pt}}*{4}{c}!{\hspace{3pt}\color{tableline}\vrule width 0.45pt\hspace{3pt}}*{5}{c}@{}}
        \arrayrulecolor{jingdongred}\toprule[0.85pt]
        \arrayrulecolor{tableline}
        & \multicolumn{4}{c}{\ablationHeading{Temporal grounding}{ablationsage}} & \multicolumn{5}{c}{\ablationHeading{Video QA}{tableclay}} \\
        \addlinespace[1.5pt]
        \multirow{-2}{*}[3pt]{{\bfseries Model}} & {\bfseries Charades\textsuperscript{TL}} & {\bfseries ActivityNet\textsuperscript{TL}} & {\bfseries QVHighlights\textsuperscript{TL}} & {\bfseries Avg.} & {\bfseries LVBench} & {\bfseries Video-MME} & {\bfseries LongVideoBench} & {\bfseries MLVU} & {\bfseries Avg.} \\
        \midrule[0.55pt]
        \tableFocusRow
        {\color{tableclay}\bfseries VideoJev-4B\textsuperscript{\ensuremath{\dagger}} (baseline)} & \ablationScore{60.6}{\textemdash}{tablemuted} & \ablationScore{{\bfseries 62.8}}{\textemdash}{tablemuted} & \ablationScore{{\bfseries 72.9}}{\textemdash}{tablemuted} & \ablationScore{{\bfseries 65.4}}{\textemdash}{tablemuted} & \ablationScore{{\bfseries 67.7}}{\textemdash}{tablemuted} & \ablationScore{{\bfseries 74.9}}{\textemdash}{tablemuted} & \ablationScore{{\bfseries 73.5}}{\textemdash}{tablemuted} & \ablationScore{{\bfseries 82.6}}{\textemdash}{tablemuted} & \ablationScore{{\bfseries 74.7}}{\textemdash}{tablemuted} \\
        Qwen3.5-4B AR SFT & \ablationScore{{\bfseries 61.0}}{(+0.4)}{ablationsage} & \ablationScore{61.3}{(-1.5)}{tableclay} & \ablationScore{70.0}{(-2.9)}{tableclay} & \ablationScore{64.1}{(-1.3)}{tableclay} & \ablationScore{66.9}{(-0.8)}{tableclay} & \ablationScore{74.5}{(-0.4)}{tableclay} & \ablationScore{71.4}{(-2.1)}{tableclay} & \ablationScore{78.6}{(-4.0)}{tableclay} & \ablationScore{72.8}{(-1.9)}{tableclay} \\
        \arrayrulecolor{tableink}\bottomrule[0.85pt]
    \end{tabular}%
    }
    \vspace{3pt}
    \endgroup
\end{table}

\paragraph{Cached inference and Joint inference.}
\emph{Cached inference} evaluates each query independently against the same video prefix, providing a reference without interference between queries. \emph{Joint inference} efficiently scores all queries in a group in one forward pass. To retain the benefits of optimized kernels such as FlashAttention, it uses standard causal attention without custom masks to isolate queries. Later queries can therefore attend to preceding questions and candidates. We verified that earlier questions do not leak answers to later ones; no reference or predicted answers are inserted. Table~\ref{tab:query-ablation} tests whether these interactions degrade prediction quality using the same trained checkpoint within each size. Surprisingly, Joint inference improves mean grounding/QA scores by 0.8/0.5 points at 4B and 0.3/0.7 points at 35B-A3B. Scores improve in 13 of the 14 model--benchmark comparisons, with only a 0.1-point decrease on QVHighlights at 35B-A3B. These small gains may reflect alignment with training, which uses the same joint layout for efficiency: the model learns to accommodate query interactions while sharing visual computation.

\begin{table}[!htb]
    \centering
    \caption{\textbf{Cached inference versus Joint inference.}}
    \label{tab:query-ablation}
    \begingroup
    \color{tableink}
    \setlength{\tabcolsep}{3.6pt}
    \renewcommand{\arraystretch}{1.16}
    \resizebox{\linewidth}{!}{%
    \fontfamily{jdlangzhengtiregular}\selectfont
    \begin{tabular}{@{}l!{\hspace{3pt}\color{tableline}\vrule width 0.45pt\hspace{3pt}}*{4}{c}!{\hspace{3pt}\color{tableline}\vrule width 0.45pt\hspace{3pt}}*{5}{c}@{}}
        \arrayrulecolor{jingdongred}\toprule[0.85pt]
        \arrayrulecolor{tableline}
        & \multicolumn{4}{c}{\ablationHeading{Temporal grounding}{ablationsage}} & \multicolumn{5}{c}{\ablationHeading{Video QA}{tableclay}} \\
        \addlinespace[1.5pt]
        \multirow{-2}{*}[3pt]{{\bfseries Setting}} & {\bfseries Charades\textsuperscript{TL}} & {\bfseries ActivityNet\textsuperscript{TL}} & {\bfseries QVHighlights\textsuperscript{TL}} & {\bfseries Avg.} & {\bfseries LVBench} & {\bfseries Video-MME} & {\bfseries LongVideoBench} & {\bfseries MLVU} & {\bfseries Avg.} \\
        \midrule[0.55pt]
        \multicolumn{10}{@{}l}{{\color{tablemuted}\bfseries VideoJev-4B}} \\
        \tableFocusRow
        {\color{tableclay}\bfseries Joint inference (baseline)} & \ablationScore{{\bfseries 60.1}}{\textemdash}{tablemuted} & \ablationScore{{\bfseries 62.8}}{\textemdash}{tablemuted} & \ablationScore{{\bfseries 73.2}}{\textemdash}{tablemuted} & \ablationScore{{\bfseries 65.3}}{\textemdash}{tablemuted} & \ablationScore{{\bfseries 67.9}}{\textemdash}{tablemuted} & \ablationScore{{\bfseries 75.0}}{\textemdash}{tablemuted} & \ablationScore{{\bfseries 73.6}}{\textemdash}{tablemuted} & \ablationScore{{\bfseries 80.6}}{\textemdash}{tablemuted} & \ablationScore{{\bfseries 74.3}}{\textemdash}{tablemuted} \\
        Cached inference & \ablationScore{58.6}{(-1.5)}{tableclay} & \ablationScore{61.8}{(-1.0)}{tableclay} & \ablationScore{73.1}{(-0.1)}{tableclay} & \ablationScore{64.5}{(-0.8)}{tableclay} & \ablationScore{67.1}{(-0.8)}{tableclay} & \ablationScore{74.9}{(-0.1)}{tableclay} & \ablationScore{72.9}{(-0.7)}{tableclay} & \ablationScore{80.1}{(-0.5)}{tableclay} & \ablationScore{73.8}{(-0.5)}{tableclay} \\
        \midrule[0.45pt]
        \multicolumn{10}{@{}l}{{\color{tablemuted}\bfseries VideoJev-35B-A3B}} \\
        \tableFocusRow
        {\color{tableclay}\bfseries Joint inference (baseline)} & \ablationScore{{\bfseries 63.7}}{\textemdash}{tablemuted} & \ablationScore{{\bfseries 65.5}}{\textemdash}{tablemuted} & \ablationScore{74.0}{\textemdash}{tablemuted} & \ablationScore{{\bfseries 67.7}}{\textemdash}{tablemuted} & \ablationScore{{\bfseries 73.4}}{\textemdash}{tablemuted} & \ablationScore{{\bfseries 81.8}}{\textemdash}{tablemuted} & \ablationScore{{\bfseries 77.1}}{\textemdash}{tablemuted} & \ablationScore{{\bfseries 85.9}}{\textemdash}{tablemuted} & \ablationScore{{\bfseries 79.5}}{\textemdash}{tablemuted} \\
        Cached inference & \ablationScore{63.1}{(-0.6)}{tableclay} & \ablationScore{65.0}{(-0.5)}{tableclay} & \ablationScore{{\bfseries 74.1}}{(+0.1)}{ablationsage} & \ablationScore{67.4}{(-0.3)}{tableclay} & \ablationScore{72.2}{(-1.2)}{tableclay} & \ablationScore{81.3}{(-0.5)}{tableclay} & \ablationScore{76.7}{(-0.4)}{tableclay} & \ablationScore{85.0}{(-0.9)}{tableclay} & \ablationScore{78.8}{(-0.7)}{tableclay} \\
        \arrayrulecolor{tableink}\bottomrule[0.85pt]
    \end{tabular}%
    }

    \endgroup
\end{table}

\paragraph{Joint training for grounding and QA.}
Temporal grounding and video QA require different forms of prediction: localizing an event in time and selecting a textual answer. We therefore examine whether the two tasks interfere, remain independent, or reinforce each other within VideoJev. Table~\ref{tab:joint-single-task} compares VideoJev-4B trained jointly on both tasks with models trained on either task alone. Joint training improves performance on all seven benchmarks, raising mean grounding mIoU by {\bfseries 0.7} points and mean QA score by {\bfseries 1.3} points, with the largest QA gain on LongVideoBench (+1.9 points). These gains indicate mutual benefit: temporal supervision may sharpen evidence localization, while QA strengthens semantic discrimination. The shared scorer provides a simple interface for learning these complementary capabilities.

\begin{table}[!htb]
    \centering
    \caption{\textbf{Joint versus single-task training.} VideoJev-4B is trained on temporal grounding (TG), video question answering (QA), or both.}
    \label{tab:joint-single-task}
    \begingroup
    \color{tableink}
    \setlength{\tabcolsep}{3.6pt}
    \renewcommand{\arraystretch}{1.16}
    \resizebox{\linewidth}{!}{%
    \fontfamily{jdlangzhengtiregular}\selectfont
    \begin{tabular}{@{}l!{\hspace{3pt}\color{tableline}\vrule width 0.45pt\hspace{3pt}}*{4}{c}!{\hspace{3pt}\color{tableline}\vrule width 0.45pt\hspace{3pt}}*{5}{c}@{}}
        \arrayrulecolor{jingdongred}\toprule[0.85pt]
        \arrayrulecolor{tableline}
        & \multicolumn{4}{c}{\ablationHeading{Temporal grounding}{ablationsage}} & \multicolumn{5}{c}{\ablationHeading{Video QA}{tableclay}} \\
        \addlinespace[1.5pt]
        \multirow{-2}{*}[3pt]{{\bfseries Setting}} & {\bfseries Charades\textsuperscript{TL}} & {\bfseries ActivityNet\textsuperscript{TL}} & {\bfseries QVHighlights\textsuperscript{TL}} & {\bfseries Avg.} & {\bfseries LVBench} & {\bfseries Video-MME} & {\bfseries LongVideoBench} & {\bfseries MLVU} & {\bfseries Avg.} \\
        \midrule[0.55pt]
        \tableFocusRow
        {\color{tableclay}\bfseries TG+QA (baseline)} & \ablationScore{{\bfseries 60.1}}{\textemdash}{tablemuted} & \ablationScore{{\bfseries 62.8}}{\textemdash}{tablemuted} & \ablationScore{{\bfseries 73.2}}{\textemdash}{tablemuted} & \ablationScore{{\bfseries 65.3}}{\textemdash}{tablemuted} & \ablationScore{{\bfseries 67.9}}{\textemdash}{tablemuted} & \ablationScore{{\bfseries 75.0}}{\textemdash}{tablemuted} & \ablationScore{{\bfseries 73.6}}{\textemdash}{tablemuted} & \ablationScore{{\bfseries 80.6}}{\textemdash}{tablemuted} & \ablationScore{{\bfseries 74.3}}{\textemdash}{tablemuted} \\
        TG-only & \ablationScore{59.6}{(-0.5)}{tableclay} & \ablationScore{61.8}{(-1.0)}{tableclay} & \ablationScore{72.5}{(-0.7)}{tableclay} & \ablationScore{64.6}{(-0.7)}{tableclay} & -- & -- & -- & -- & -- \\
        QA-only & -- & -- & -- & -- & \ablationScore{66.3}{(-1.6)}{tableclay} & \ablationScore{74.9}{(-0.1)}{tableclay} & \ablationScore{71.7}{(-1.9)}{tableclay} & \ablationScore{79.2}{(-1.4)}{tableclay} & \ablationScore{73.0}{(-1.3)}{tableclay} \\
        \arrayrulecolor{tableink}\bottomrule[0.85pt]
    \end{tabular}%
    }

    \endgroup
\end{table}

\paragraph{Auxiliary supervision.}
Correct local judgments must also cover the event as a whole. Table~\ref{tab:objective-ablation} tests the value of adding Soft IoU to the grounding objective in Eq.~\eqref{eq:joint-grounding-loss}, while QA retains cross-entropy. Soft IoU raises mean grounding mIoU from 60.7 to 65.3 ({\bfseries +4.6} points) and mean QA score from 73.9 to 74.3 ({\bfseries +0.4} points), with improvements on all seven benchmarks. These results highlight the value of event-level coverage supervision and suggest that better temporal learning also benefits the shared representations used for QA.

\begin{table}[!htb]
    \centering
    \caption{\textbf{Effect of Soft IoU supervision in VideoJev-4B.} Soft IoU augments the grounding BCE objective; QA retains cross-entropy. Deltas are relative to BCE/CE only (baseline). Both variants use two-stage training.}
    \label{tab:objective-ablation}
    \begingroup
    \color{tableink}
    \setlength{\tabcolsep}{3.6pt}
    \renewcommand{\arraystretch}{1.16}
    \resizebox{\linewidth}{!}{%
    \fontfamily{jdlangzhengtiregular}\selectfont
    \begin{tabular}{@{}l!{\hspace{3pt}\color{tableline}\vrule width 0.45pt\hspace{3pt}}*{4}{c}!{\hspace{3pt}\color{tableline}\vrule width 0.45pt\hspace{3pt}}*{5}{c}@{}}
        \arrayrulecolor{jingdongred}\toprule[0.85pt]
        \arrayrulecolor{tableline}
        & \multicolumn{4}{c}{\ablationHeading{Temporal grounding}{ablationsage}} & \multicolumn{5}{c}{\ablationHeading{Video QA}{tableclay}} \\
        \addlinespace[1.5pt]
        \multirow{-2}{*}[3pt]{{\bfseries Setting}} & {\bfseries Charades\textsuperscript{TL}} & {\bfseries ActivityNet\textsuperscript{TL}} & {\bfseries QVHighlights\textsuperscript{TL}} & {\bfseries Avg.} & {\bfseries LVBench} & {\bfseries Video-MME} & {\bfseries LongVideoBench} & {\bfseries MLVU} & {\bfseries Avg.} \\
        \midrule[0.55pt]
        \tableFocusRow
        {\color{tableclay}\bfseries BCE/CE only (baseline)} & \ablationScore{53.3}{\textemdash}{tablemuted} & \ablationScore{58.8}{\textemdash}{tablemuted} & \ablationScore{70.0}{\textemdash}{tablemuted} & \ablationScore{60.7}{\textemdash}{tablemuted} & \ablationScore{67.4}{\textemdash}{tablemuted} & \ablationScore{74.6}{\textemdash}{tablemuted} & \ablationScore{72.9}{\textemdash}{tablemuted} & \ablationScore{80.5}{\textemdash}{tablemuted} & \ablationScore{73.9}{\textemdash}{tablemuted} \\
        \quad + Soft IoU & \ablationScore{{\bfseries 60.1}}{(+6.8)}{ablationsage} & \ablationScore{{\bfseries 62.8}}{(+4.0)}{ablationsage} & \ablationScore{{\bfseries 73.2}}{(+3.2)}{ablationsage} & \ablationScore{{\bfseries 65.3}}{(+4.6)}{ablationsage} & \ablationScore{{\bfseries 67.9}}{(+0.5)}{ablationsage} & \ablationScore{{\bfseries 75.0}}{(+0.4)}{ablationsage} & \ablationScore{{\bfseries 73.6}}{(+0.7)}{ablationsage} & \ablationScore{{\bfseries 80.6}}{(+0.1)}{ablationsage} & \ablationScore{{\bfseries 74.3}}{(+0.4)}{ablationsage} \\
        \arrayrulecolor{tableink}\bottomrule[0.85pt]
    \end{tabular}%
    }

    \endgroup
\end{table}

\paragraph{Final-layer MLP removal.}
The final MLP adds a transformation before candidate scoring, but Table~\ref{tab:mlp-ablation} shows that it is not needed to maintain prediction quality. Removing it raises mean grounding and QA scores by 0.2 points each, leaving overall performance essentially unchanged. Following Occam's razor, we adopt the simpler readout: pre-MLP features preserve both capabilities while avoiding an additional layer of computation.

\begin{table}[!htb]
    \centering
    \caption{\textbf{Effect of the final-layer MLP in VideoJev-4B.}}
    \label{tab:mlp-ablation}
    \begingroup
    \color{tableink}
    \setlength{\tabcolsep}{3.6pt}
    \renewcommand{\arraystretch}{1.16}
    \resizebox{\linewidth}{!}{%
    \fontfamily{jdlangzhengtiregular}\selectfont
    \begin{tabular}{@{}l!{\hspace{3pt}\color{tableline}\vrule width 0.45pt\hspace{3pt}}*{4}{c}!{\hspace{3pt}\color{tableline}\vrule width 0.45pt\hspace{3pt}}*{5}{c}@{}}
        \arrayrulecolor{jingdongred}\toprule[0.85pt]
        \arrayrulecolor{tableline}
        & \multicolumn{4}{c}{\ablationHeading{Temporal grounding}{ablationsage}} & \multicolumn{5}{c}{\ablationHeading{Video QA}{tableclay}} \\
        \addlinespace[1.5pt]
        \multirow{-2}{*}[3pt]{{\bfseries Setting}} & {\bfseries Charades\textsuperscript{TL}} & {\bfseries ActivityNet\textsuperscript{TL}} & {\bfseries QVHighlights\textsuperscript{TL}} & {\bfseries Avg.} & {\bfseries LVBench} & {\bfseries Video-MME} & {\bfseries LongVideoBench} & {\bfseries MLVU} & {\bfseries Avg.} \\
        \midrule[0.55pt]
        \tableFocusRow
        {\color{tableclay}\bfseries Remove final MLP (baseline)} & \ablationScore{60.1}{\textemdash}{tablemuted} & \ablationScore{{\bfseries 62.8}}{\textemdash}{tablemuted} & \ablationScore{{\bfseries 73.2}}{\textemdash}{tablemuted} & \ablationScore{{\bfseries 65.3}}{\textemdash}{tablemuted} & \ablationScore{{\bfseries 67.9}}{\textemdash}{tablemuted} & \ablationScore{{\bfseries 75.0}}{\textemdash}{tablemuted} & \ablationScore{{\bfseries 73.6}}{\textemdash}{tablemuted} & \ablationScore{80.6}{\textemdash}{tablemuted} & \ablationScore{{\bfseries 74.3}}{\textemdash}{tablemuted} \\
        Retain final MLP & \ablationScore{{\bfseries 60.5}}{(+0.4)}{ablationsage} & \ablationScore{62.1}{(-0.7)}{tableclay} & \ablationScore{72.7}{(-0.5)}{tableclay} & \ablationScore{65.1}{(-0.2)}{tableclay} & \ablationScore{{\bfseries 67.9}}{(0.0)}{tablemuted} & \ablationScore{74.7}{(-0.3)}{tableclay} & \ablationScore{73.0}{(-0.6)}{tableclay} & \ablationScore{{\bfseries 80.8}}{(+0.2)}{ablationsage} & \ablationScore{74.1}{(-0.2)}{tableclay} \\
        \arrayrulecolor{tableink}\bottomrule[0.85pt]
    \end{tabular}%
    }

    \endgroup
\end{table}

\paragraph{Two-stage adaptation.}
The two-stage schedule first aligns the scorer with frozen features, then adapts the language backbone and scorer together. Table~\ref{tab:schedule-ablation} shows broadly comparable performance to single-stage training: two-stage adaptation scores slightly higher on four of seven benchmarks, while single-stage training leads by only 0.1 point in mean grounding mIoU and 0.4 points in mean QA, mainly due to MLVU. Two-stage adaptation achieves this quality while updating the language backbone on only 70\% of the videos; the remaining 30\% use cheaper head-only training. We adopt this schedule for its lower cost and comparable performance; aligning the scorer first may also stabilize backbone adaptation.

\begin{table}[!htb]
    \centering
    \caption{\textbf{Two-stage versus single-stage adaptation of VideoJev-4B.} The baseline warms up the head on 30\% of videos and jointly adapts on the remaining 70\%; single-stage training directly adapts on all videos with the full objectives.}
    \label{tab:schedule-ablation}
    \begingroup
    \color{tableink}
    \setlength{\tabcolsep}{3.6pt}
    \renewcommand{\arraystretch}{1.16}
    \resizebox{\linewidth}{!}{%
    \fontfamily{jdlangzhengtiregular}\selectfont
    \begin{tabular}{@{}l!{\hspace{3pt}\color{tableline}\vrule width 0.45pt\hspace{3pt}}*{4}{c}!{\hspace{3pt}\color{tableline}\vrule width 0.45pt\hspace{3pt}}*{5}{c}@{}}
        \arrayrulecolor{jingdongred}\toprule[0.85pt]
        \arrayrulecolor{tableline}
        & \multicolumn{4}{c}{\ablationHeading{Temporal grounding}{ablationsage}} & \multicolumn{5}{c}{\ablationHeading{Video QA}{tableclay}} \\
        \addlinespace[1.5pt]
        \multirow{-2}{*}[3pt]{{\bfseries Setting}} & {\bfseries Charades\textsuperscript{TL}} & {\bfseries ActivityNet\textsuperscript{TL}} & {\bfseries QVHighlights\textsuperscript{TL}} & {\bfseries Avg.} & {\bfseries LVBench} & {\bfseries Video-MME} & {\bfseries LongVideoBench} & {\bfseries MLVU} & {\bfseries Avg.} \\
        \midrule[0.55pt]
        \tableFocusRow
        {\color{tableclay}\bfseries Two-stage (baseline)} & \ablationScore{60.1}{\textemdash}{tablemuted} & \ablationScore{{\bfseries 62.8}}{\textemdash}{tablemuted} & \ablationScore{{\bfseries 73.2}}{\textemdash}{tablemuted} & \ablationScore{65.3}{\textemdash}{tablemuted} & \ablationScore{{\bfseries 67.9}}{\textemdash}{tablemuted} & \ablationScore{{\bfseries 75.0}}{\textemdash}{tablemuted} & \ablationScore{{\bfseries 73.6}}{\textemdash}{tablemuted} & \ablationScore{80.6}{\textemdash}{tablemuted} & \ablationScore{74.3}{\textemdash}{tablemuted} \\
        Single-stage (100\%) & \ablationScore{{\bfseries 60.6}}{(+0.5)}{ablationsage} & \ablationScore{{\bfseries 62.8}}{(0.0)}{tablemuted} & \ablationScore{72.9}{(-0.3)}{tableclay} & \ablationScore{{\bfseries 65.4}}{(+0.1)}{ablationsage} & \ablationScore{67.7}{(-0.2)}{tableclay} & \ablationScore{74.9}{(-0.1)}{tableclay} & \ablationScore{73.5}{(-0.1)}{tableclay} & \ablationScore{{\bfseries 82.6}}{(+2.0)}{ablationsage} & \ablationScore{{\bfseries 74.7}}{(+0.4)}{ablationsage} \\
        \arrayrulecolor{tableink}\bottomrule[0.85pt]
    \end{tabular}%
    }

    \endgroup
\end{table}

\endgroup

% Keep headings with their opening lines; let figures float so prose fills the page.
\Needspace{4\baselineskip}
\subsection{Analysis}
\label{sec:analysis}

\begin{figure}[!htb]
    \centering
    \includegraphics[width=\linewidth]{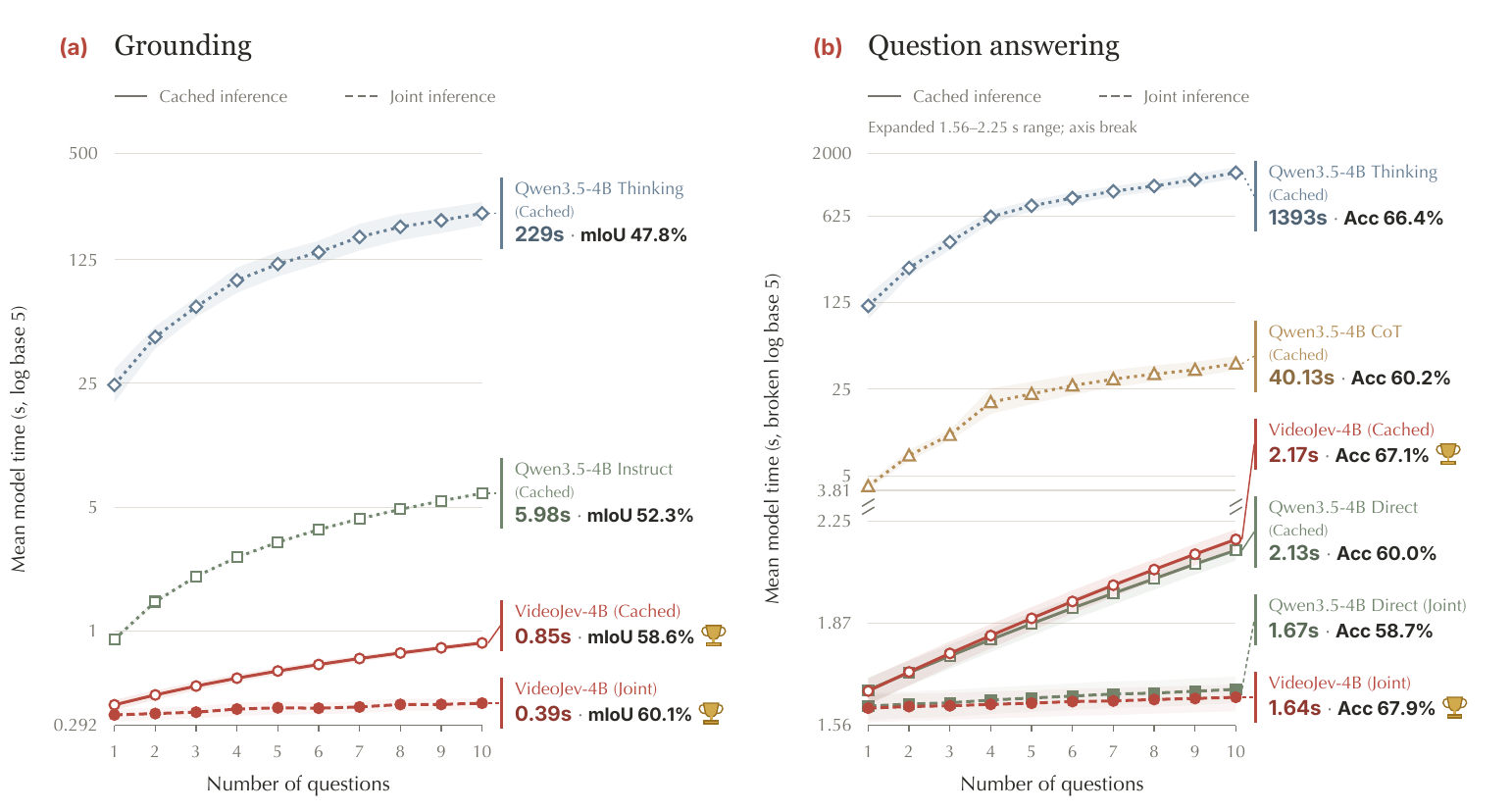}
    \caption{\textbf{Multi-query efficiency at 4B.} Mean forward time for 1--10 queries per video on Charades (a) and Video-Holmes~\citep{cheng2025video} (b); shading shows 95\% confidence intervals. Solid/dotted lines use \emph{Cached inference} (dotted: multi-token generation); dashed lines use \emph{Joint inference}. End labels pair ten-query latency with separately evaluated full-benchmark Charades mIoU or LVBench accuracy; quality settings follow the source evaluations in Appendix~\ref{app:performance}.}
    \label{fig:query-efficiency}
\end{figure}

\paragraph{Query scaling and inference efficiency.}
Figure~\ref{fig:query-efficiency} compares the computational efficiency of \emph{Cached inference} and \emph{Joint inference} as the number of queries about the same video increases. VideoJev supports both modes. Autoregressive QA supports joint processing in \emph{Direct} mode, which predicts exactly one option token without reasoning. Multi-token outputs, including temporal intervals and reasoning traces, require sequential token feedback and use \emph{Cached inference}. Appendix~\ref{app:inference-modes} details these modes.

We measure 1--10 queries about the same video at batch size 1, counting visual encoding and prefix prefill once. For each task, we randomly sample 40 videos within duration strata. End labels report full-benchmark Charades mIoU and LVBench accuracy. Quality is evaluated separately from timing; Appendices~\ref{app:latency} and~\ref{app:performance} specify the respective protocols and remaining score sources.

VideoJev's \emph{Joint inference} adds only {\bfseries 55\,ms} for grounding and {\bfseries 31\,ms} for QA as query count grows from one to ten. At ten queries, forward times are 0.39\,s and 1.64\,s, versus 0.85\,s and 2.17\,s under \emph{Cached inference}, while independently measured Charades mIoU and LVBench accuracy improve by 1.5 and 0.8 points, respectively. The autoregressive Direct QA baseline benefits less from joint processing: its accuracy drops from 60.0\% under \emph{Cached inference} to 58.7\% under \emph{Joint inference}, below VideoJev's 67.9\% at comparable latency (1.67\,s versus 1.64\,s). Multi-token generation remains costly: Qwen3.5-4B Instruct takes 5.98\,s for ten grounding queries at 52.3 mIoU, versus VideoJev's 0.39\,s and 60.1 mIoU; Qwen Thinking takes 1,393\,s for ten QA queries at 66.4\% accuracy. The small marginal cost comes without a loss in prediction quality.

\begin{figure}[!htb]
    \centering
    \includegraphics[width=\linewidth]{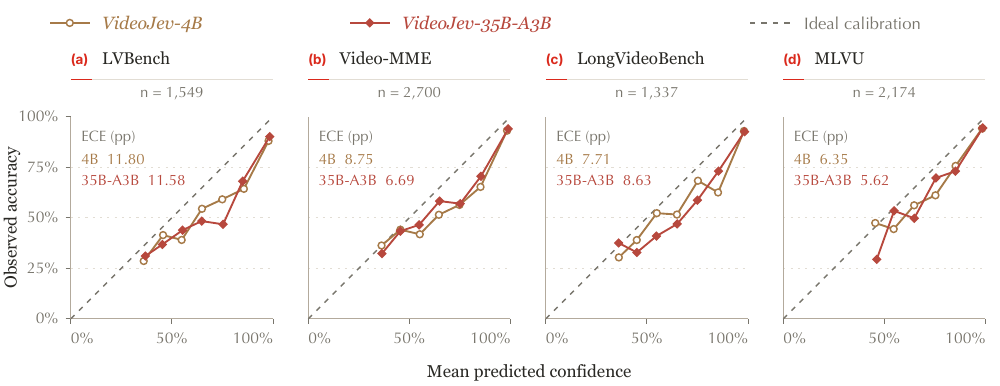}
    \caption{\textbf{Confidence calibration on Video QA.} Points show mean confidence and observed accuracy in equal-width bins containing at least 20 questions; all bins contribute to ECE. The dashed diagonal denotes perfect calibration. MLVU uses seven-task-balanced weights; the other benchmarks weight questions equally. Curves use the reported per-benchmark predictions.}
    \label{fig:qa-calibration}
\end{figure}

\paragraph{Confidence and calibration.}
Candidate scores also provide a signal for deciding when to trust a prediction. We examine the highest option probability, $\max_i p_i$, using the predictions from Table~\ref{tab:video-qa}. Higher confidence generally accompanies higher accuracy (Figure~\ref{fig:qa-calibration}). On LVBench, the 35B-A3B model reaches 95.8\% accuracy on its 591 predictions with confidence at least 99\%, compared with 73.4\% overall. This separation suggests potential for selecting reliable answers or escalating uncertain cases.

Calibration requires confidence to match the frequency of correct predictions~\citep{gneiting2007calibration,guo2017calibration}. Jev explicitly optimizes this property through reinforcement learning~\citep{typesafe2026jev}; VideoJev receives no calibration-specific training. Its probabilities tend to overestimate correctness: the four-benchmark mean expected calibration error (ECE; ten equal-width bins) is 8.65 and 8.13 percentage points for 4B and 35B-A3B. Temperature scaling on an independent calibration set~\citep{guo2017calibration} or calibration-aware training could address this gap.

% Keep analysis figures ahead of Related Work.
\FloatBarrier

% !TEX root = ../main.tex
\section{Related Work}

\paragraph{MLLM video understanding.}
VideoChat~\citep{li2023videochat} and Video-LLaMA~\citep{zhang2023videollama} connect visual representations to language models for instruction-driven video understanding. VideoChat2~\citep{li2024mvbenchcomprehensivemultimodalvideo} and VideoChat3~\citep{li2026videochat3fullyopenvideo} expand these capabilities through multimodal training and diverse supervision. For long videos, VideoChat-Flash~\citep{li2025videochatflash} compresses visual tokens, while Video-o3~\citep{zeng2026videoo3} interleaves reasoning with visual clue seeking. VideoLLM-online~\citep{chen2024videollmonline} supports streaming interaction with continuous cache reuse, and StreamingVLM~\citep{xu2026streamingvlm} maintains a bounded cache for long-running video streams. JoyAI-VL-Interaction~\citep{yao2026joyaivlinteraction} learns when to remain silent, respond, or delegate during real-time interaction, while OneStreamer~\citep{zeng2026onestreamerunifyingperceptionmemory} jointly models streaming perception, memory formation, and proactive response. Complementary work studies vision-language transfer in AWT~\citep{zhu2024awt} and efficient test-time prompt adaptation~\citep{zhu2024selftpt}. Benchmarks probe compositional understanding in TVQA~\citep{lei-etal-2018-tvqa}, temporal and causal reasoning in NExT-QA~\citep{xiao2021nextqanextphasequestionansweringexplaining}, long-horizon understanding in EgoSchema~\citep{mangalam2023egoschemadiagnosticbenchmarklongform}, and broader video skills in MVBench and Video-MME~\citep{li2024mvbenchcomprehensivemultimodalvideo,fu2025videommefirstevercomprehensiveevaluation}. VideoJev focuses on direct decisions over time and options, sharing an encoded video context across multiple queries.

\paragraph{Temporal grounding.}
Temporal grounding localizes language-described events through candidate matching~\citep{Gao_2017_ICCV,zhang2020learning}, span prediction~\citep{zhang-etal-2020-span}, moment-set prediction~\citep{lei2021neurips-detecting}, or clip-level foreground, boundary, and saliency outputs~\citep{Lin_2023_ICCV}. In temporal action detection, DualDETR~\citep{zhu2024dualdetr} jointly refines instance and boundary queries. TimeChat and VTimeLLM bring temporal grounding into video MLLMs~\citep{ren2024timechattimesensitivemultimodallarge,huang2023vtimellmempowerllmgrasp}; LITA and Grounded-VideoLLM introduce explicit temporal tokens~\citep{huang2024lita,wang2025groundedvideollm}. TimeSuite uses grounded tuning to improve long-video understanding~\citep{zeng2025timesuite}, and TimeLens2 advances generalist temporal modeling~\citep{zhu2026timelens2generalistvideotemporal}. DisTime predicts boundary distributions~\citep{zeng2025distime}, while controlled comparisons examine temporal output designs~\citep{jin2026temporalparadigms}. Recent analysis also extracts temporal evidence from prefill attention to guide subsequent answer generation~\citep{du2026knowwhen}. Grounding and QA have already been combined: GroundVQA uses a temporal localizer and an answer decoder~\citep{di2024groundvqa}, while SEAL learns compact representations for both tasks~\citep{wang2025seal}. VideoJev uses a shared scoring head for temporal membership and textual options, decoding a contiguous span from temporal scores.

\paragraph{System One models.}
Jev~\citep{typesafe2026jev,typesafe2026systemone} introduces typed probabilistic judgments, while this-that-model-1.0 and JevLite explore one-pass decision readouts~\citep{cheng2026thisthatmodel10typeddecisionmodel,ren2026openjevjudgmentscallscreenbenchcalibrated}. Visual Jev~\citep{yu2026visualjevaccurateefficient} evaluates multiple forced-choice questions over a shared image, and PixelJev~\citep{zhou2026textdecisionspixelsstudy} supports image recognition and multiple-choice VQA. FreeRet~\citep{zhu2025freeret} extracts multimodal representations and reranks candidates using an MLLM's multiple-choice probabilities; its reranker can be viewed as a training-free route to Jev-style relevance judgments. Related video studies assess quality from metadata and derived features~\citep{robitza2026jevqavideoquality} or anomalies from captions and summaries~\citep{qin2026decisionreadoutstextmediatedvideo}. VideoJev extends this paradigm to temporal decisions and repeated video queries, aligning multi-query training with inference.

% !TEX root = ../main.tex
\section{Future Work}
\label{sec:future-work}

\paragraph{Toward general-purpose visual decisions.}
Beyond temporal grounding and multiple-choice Video QA, tasks such as action recognition, event detection, scene-state classification, and visual condition verification can also be formulated as decisions over candidate hypotheses. Extending the candidate interface to these tasks and output structures could reduce reliance on task-specific models and prediction heads.

\paragraph{From closed-set decisions to open-ended agent workflows.}
A higher-level agent could decompose open-ended goals into visual checks using standard operating procedures or generated plans. It could propose candidate events, actions, or states for VideoJev to evaluate, then use these judgments to guide planning and execution. Ambiguous or unexpected observations could trigger candidate revision or further reasoning. Here, \emph{System Two organizes the decision space, while System One evaluates it against visual evidence}. Applications include procedural monitoring, interactive assistance, embodied agents, and computer-use systems.

\paragraph{Toward streaming and reliable visual interaction.}
Streaming inputs will require incremental visual encoding, efficient memory management, and timely updates to the decision context. Reliability also needs improvement: higher confidence generally accompanies higher Video QA accuracy, but predicted probabilities tend to overestimate correctness. Calibration-aware training, uncertainty estimation, and selective prediction could help determine when to accept System One decisions or defer to higher-level reasoning.

% !TEX root = ../main.tex
\section{Conclusion}

We introduced VideoJev, a family of video System One models that unifies temporal grounding and multiple-choice video QA through direct candidate scoring. \emph{Cached inference} and \emph{Joint inference} make repeated judgments efficient by reusing visual computation. Across seven benchmarks, dense 4B and MoE 35B-A3B models achieve strong prediction quality and inference efficiency. Ablations show that these gains extend beyond training data alone and that the two tasks benefit from joint learning. VideoJev thus offers a foundation for visual agents that compose structured judgments into complex, open-ended workflows.

\bibliographystyle{assets/plainnat}
\bibliography{references}
% !TEX root = ../main.tex
\clearpage
\appendix
\section*{Supplementary Material}
\subsection*{Contents}
\noindent\hyperref[app:protocol]{\ref*{app:protocol}\quad Detailed Evaluation Protocol}\dotfill\pageref*{app:protocol}\par
\noindent\hspace*{1em}\hyperref[app:latency]{\ref*{app:latency}\quad Latency}\dotfill\pageref*{app:latency}\par
\noindent\hspace*{1em}\hyperref[app:performance]{\ref*{app:performance}\quad Performance}\dotfill\pageref*{app:performance}\par
\noindent\hspace*{1em}\hyperref[app:same-data-control]{\ref*{app:same-data-control}\quad Autoregressive Control with the Same Training Data}\dotfill\pageref*{app:same-data-control}\par
\noindent\hspace*{1em}\hyperref[app:inference-modes]{\ref*{app:inference-modes}\quad Inference Modes and Direct Baseline}\dotfill\pageref*{app:inference-modes}\par
\noindent\hyperref[app:qualitative]{\ref*{app:qualitative}\quad Qualitative Benchmark Cases}\dotfill\pageref*{app:qualitative}\par
\medskip

\section{Detailed Evaluation Protocol}
\label{app:protocol}
Generative baselines use Instruct unless labeled Thinking; VideoJev defaults to \emph{Joint inference}. Both tasks use video frames without subtitles or audio. Main benchmarks use full-video sampling at 4 fps for grounding and up to 2 fps (at most 2,048 frames) for QA, with frame area capped at $640\times640$ pixels.

\subsection{Latency}
\label{app:latency}
Both tasks use initially idle H200 GPUs at batch size 1. We sum synchronized forward time for visual encoding, prefill, and all decoding steps through EOS or the model-specific output limit, including Thinking tokens, then divide by the query count. Video loading and decoding, preprocessing, transfers, and postprocessing (including parsing and metric computation) are excluded; truncated and invalid outputs remain included.

VideoJev encodes each video once and scores its queries in one forward. Generative baselines reuse visual encoding and prefix prefill, then decode from independent query branches without answer history.

\paragraph{Temporal grounding latency.}
Table~\ref{tab:temporal-grounding} covers all 1,313 Charades videos (3,363 queries) on a single H200, with one untimed warm-up per video. VideoJev removes the final-layer MLP, retains native FP32 scoring heads and MoE routers, and matches the visual inputs of Qwen3.5-4B and Video-ORA-4B. Greedy decoding allows 4,096 output tokens (Nemotron: 1,024). Context is 262,144 tokens, except for Molmo2 (110,592; YaRN factor 3). Mage lowers frame resolution only to avoid context overflow; native visual merging and pruning are retained.

\paragraph{Video QA latency.}
Table~\ref{tab:video-qa} covers all 103 LVBench test videos (1,549 questions) on a single H200, measuring each video once after untimed warm-up. VideoJev-4B and Qwen use fixed RGB geometry with the QA sampling defaults, a 256K context, and an 80\% visual-token budget; other architectures retain native visual settings and backends. Instruct requests an option letter and Thinking an answer field; Qwen3.5-4B Thinking retains its earlier prompt. VideoJev-35B-A3B retains FP32 routers and a FP32 decision head. Qwen3.5/3.6 MoE and VideoJev-35B-A3B use Transformers' native \texttt{grouped\_mm} experts.

\paragraph{Multi-query scaling.}
Figure~\ref{fig:query-efficiency} uniformly samples 40 distinct videos per task with a fixed seed: ten per Charades duration quartile and five per Video-Holmes~\citep{cheng2025video} duration stratum (eight strata). All methods use the same ordered 1--10 queries per video, cycling native questions when needed, at 2 fps, at most $640\times640$ pixels per frame, and a 262,144-token context. Inference modes follow Appendix~\ref{app:inference-modes}; Qwen sampling follows Appendix~\ref{app:performance}, with at most 32,768 output tokens. Shared visual encoding and prefill are charged once, followed by query and decoding forwards. On exclusively idle H200 GPUs, we run an untimed warm-up and three fixed-seed traces per video, retaining the whole trace with median ten-query time. We report equal-video means and pointwise 95\% confidence intervals from 10,000 paired bootstrap resamples within duration strata. Generation monitoring overhead is measured separately and excluded.

\subsection{Performance}
\label{app:performance}
\paragraph{Temporal grounding performance.}
Qwen3.5-4B/35B-A3B, Qwen3.6-35B-A3B, Mage-VL, and Nemotron are evaluated on all Charades (3,363), ActivityNet (4,500), and QVHighlights (1,541) test queries. Full videos use 4 fps, reducing frame area to fit a 144K (147,456-token) context and 114,688-token input budget. Decoding follows Appendix~\ref{app:latency}; invalid intervals count as failures. Other baselines retain published scores. VideoJev uses the same budget with \emph{Joint inference} and fixes $\tau=0.25$ and smoothing window 1 across both sizes and all experiments.

\paragraph{Video QA performance.}
All VideoJev QA evaluations---both sizes, inference modes, and ablations---use the QA sampling defaults and a fixed 256K (262,144-token) context. Frame area is reduced to keep visual tokens within 80\% of this budget. MLVU averages seven development tasks.

Generative baselines answer questions independently without answer history, unlike the shared-prefix latency protocol. Qwen also uses the QA sampling defaults and a 256K token budget. Qwen3.5/3.6 Instruct requests one option letter; Thinking retains the earlier answer-field prompt. Nemotron retains Conv3D temporal merging with factor 2 and disables EVS, with Instruct and Thinking evaluated separately. Published scores retain their source settings. Before scoring, we verify complete prediction and answer-extraction coverage. Deterministic option parsing is followed, when needed, by a text-only extractor without ground-truth access; invalid answers count as incorrect.

\paragraph{Multi-query prediction quality.}
Figure~\ref{fig:query-efficiency} reports full-Charades mIoU and full-LVBench accuracy, evaluated separately from timing. All grounding scores, including the fresh Qwen Thinking evaluation, follow the grounding input protocol above. Direct predicts exactly one option token and is evaluated separately from the Instruct baseline in Table~\ref{tab:video-qa}.
Qwen uses temperature/top-$p$ 0.7/0.8 (Instruct) or 1.0/0.95 (Thinking), top-$k$ 20, presence penalty 1.5, and at most 32,768 output tokens. Direct masks invalid option tokens before sampling filters; CoT reasons before emitting \texttt{Answer: X}. Scoring accepts unambiguous final options with optional answer text, ignores preceding reasoning, and assigns zero to invalid or truncated answers.

\Needspace{6\baselineskip}
\subsection{Autoregressive Control with the Same Training Data}
\label{app:same-data-control}
Table~\ref{tab:same-data-control} compares single-stage VideoJev-4B (Table~\ref{tab:schedule-ablation}) with native Qwen3.5-4B retaining its final MLP and tied language-model head. Both train on the same 404,037 questions from 58,424 videos across VideoJev's training partitions. The control trains for one epoch (546 updates) with XTuner FSDP2, global batch size 64, 147,456-token packs, a $5\times10^{-6}$ learning rate, 5\% warm-up, and cosine decay to $5\times10^{-7}$. Inputs and optimizer settings follow VideoJev's joint adaptation; the video encoder and merger stay frozen. Answer-token cross-entropy includes EOS and is averaged within questions, then across questions. Targets are all annotated intervals as JSON pairs in seconds, or a QA option letter; Thinking is disabled.

Evaluation uses VideoJev's visual features, full benchmark splits, and context budgets: 147,456 tokens for grounding and 262,144 for QA. Queries branch independently from an immutable video prefix without answer history. Greedy decoding allows 256 tokens for interval lists and 16 for QA. Grounding scores the first interval, counting invalid or truncated outputs as failures; QA uses benchmark-specific prompts and deterministic option parsing.

\begin{figure}[!t]
    \centering
    \includegraphics[width=\linewidth]{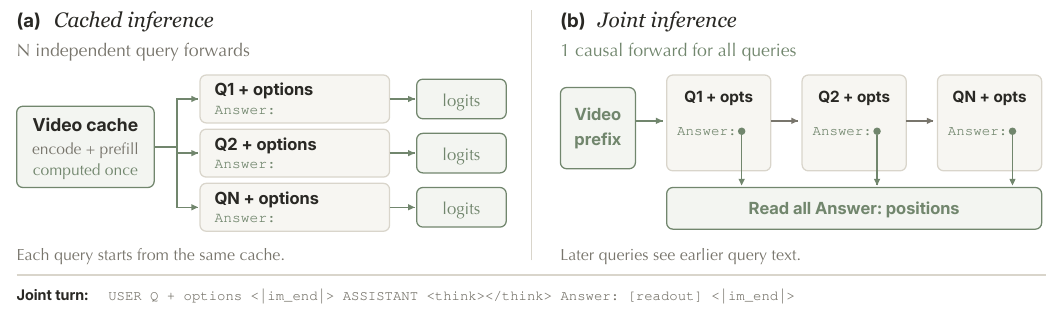}
    \caption{\textbf{Qwen3.5-4B Direct inference.} (a) Queries branch independently from a shared video cache. (b) All option predictions are read in one causal forward. Both modes use batch size 1, restrict logits to valid option tokens, and leave answer slots empty; \texttt{[readout]} marks a logit readout, not inserted text.}
    \label{fig:qwen-direct}
\end{figure}

\Needspace{8\baselineskip}
\subsection{Inference Modes and Direct Baseline}
\label{app:inference-modes}
\label{app:direct-baseline}
\label{app:empty-answer-history}
\paragraph{Shared inference modes.}
\emph{Cached inference} encodes and prefills the video once, then processes queries sequentially from independent copies of the prefix cache. Each branch contains only its query and any generated tokens, without query or answer history. \emph{Joint inference} concatenates queries after one video prefix and reads all predictions in one causal forward. Later queries can attend to earlier queries, but no reference or predicted answers are inserted. Both modes use batch size 1.

VideoJev supports both modes for grounding and QA through direct candidate scores. Autoregressive \emph{Joint inference} requires single-token answers read from each query's next-token distribution. Multi-token outputs, including intervals, CoT, and Thinking, require sequential token feedback and thus use \emph{Cached inference}. Their dotted curves in Figure~\ref{fig:query-efficiency} use this mode even when all queries are available together.

\paragraph{Direct option prediction.}
Direct predicts exactly one option token without reasoning. Figure~\ref{fig:qwen-direct} shows Qwen3.5-4B's native non-thinking template. Cached reads option logits after one query suffix forward; Joint reads every \texttt{Answer:} position in one forward over complete turns with empty answer slots. Both mask logits to valid plain or leading-space option tokens before applying official sampling filters.

\paragraph{Empty answer histories.}
With per-turn non-thinking prefixes, unconstrained LVBench readouts produced valid options for all 103 first questions but \texttt{<|im\_end|>} for 1,418 of 1,446 later questions. Inserting earlier answers, even incorrect ones, restored option outputs, suggesting sensitivity to empty answer histories. Reported Direct results use complete turns, uniform official non-thinking prefixes, no inserted answers, and the option-token masking above; unconstrained outputs are archived separately.

% Individual cases can follow the protocol without leaving a partly empty page.
\begingroup
\setlength{\intextsep}{5pt}
\section{Qualitative Benchmark Cases}
\label{app:qualitative}
Eight cases across seven benchmarks compare \emph{Joint inference} outputs. Frames retain native aspect ratios; details are enlarged for display only. Green denotes ground truth; brown and red denote the models. QA bars show uncalibrated scores with selected options outlined. Grounding times are video-relative.

\begin{figure}[!ht]
\centering
\includegraphics[width=\linewidth,trim=0 3bp 0 6bp,clip]{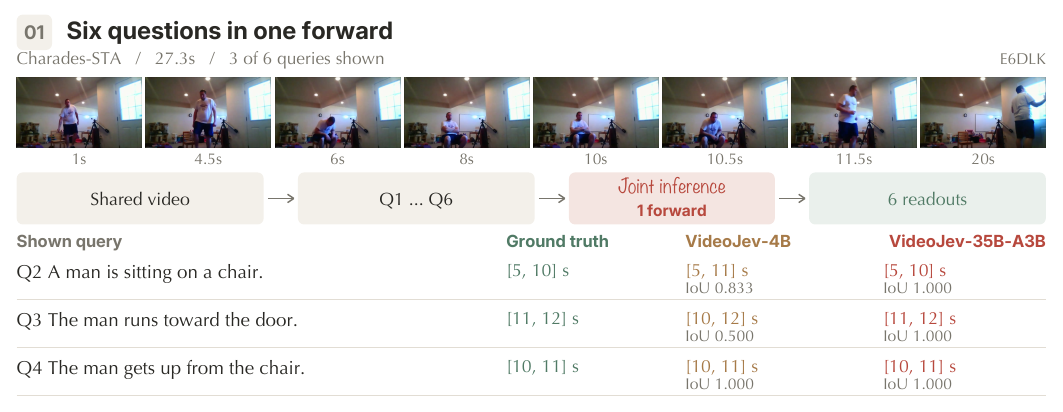}
\caption{Joint inference grounds six Charades queries in one forward; three readouts are shown in order. Both models match the one-second transition, while VideoJev-35B-A3B locates adjacent actions more precisely.}
\label{fig:qualitative-01}
\end{figure}

\clearpage
\begin{figure}[!ht]
\centering
\includegraphics[width=\linewidth,trim=0 3bp 0 6bp,clip]{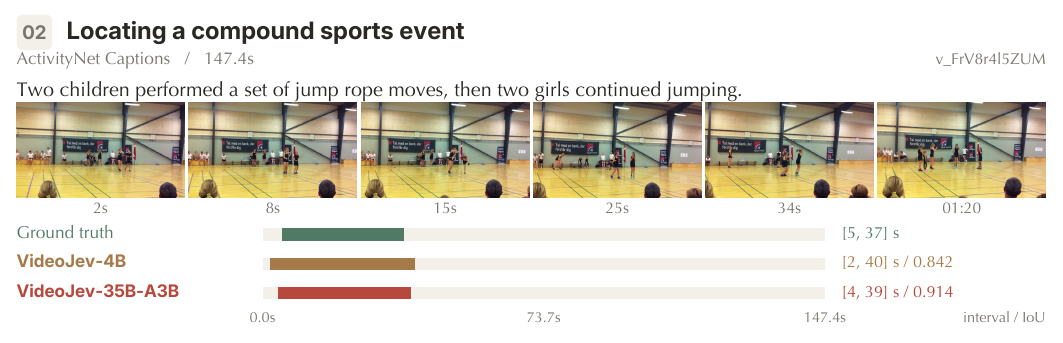}
\caption{Localizing a compound jump-rope event on ActivityNet Captions. VideoJev-35B-A3B more closely matches the annotated interval.}
\label{fig:qualitative-02}
\end{figure}

\begin{figure}[!ht]
\centering
\includegraphics[width=\linewidth,trim=0 3bp 0 6bp,clip]{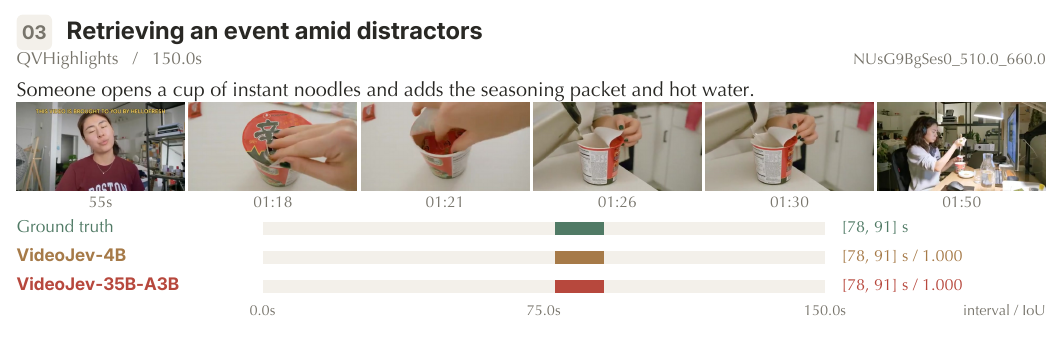}
\caption{Retrieving a preparation sequence amid vlog distractors on QVHighlights. Both models match the annotated interval.}
\label{fig:qualitative-03}
\end{figure}

\begin{figure}[!ht]
\centering
\includegraphics[width=\linewidth,trim=0 3bp 0 6bp,clip]{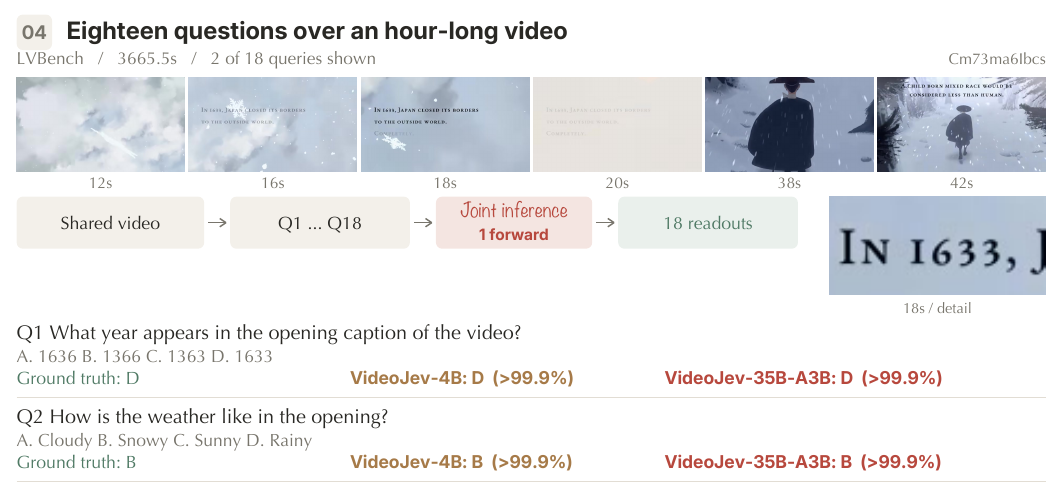}
\caption{Joint inference answers 18 questions over a 3,665.5-second LVBench video in one forward. Two readouts test caption retrieval and weather recognition; the caption detail is enlarged for display.}
\label{fig:qualitative-04}
\end{figure}

\clearpage
\begin{figure}[!ht]
\centering
\includegraphics[width=\linewidth,trim=0 3bp 0 6bp,clip]{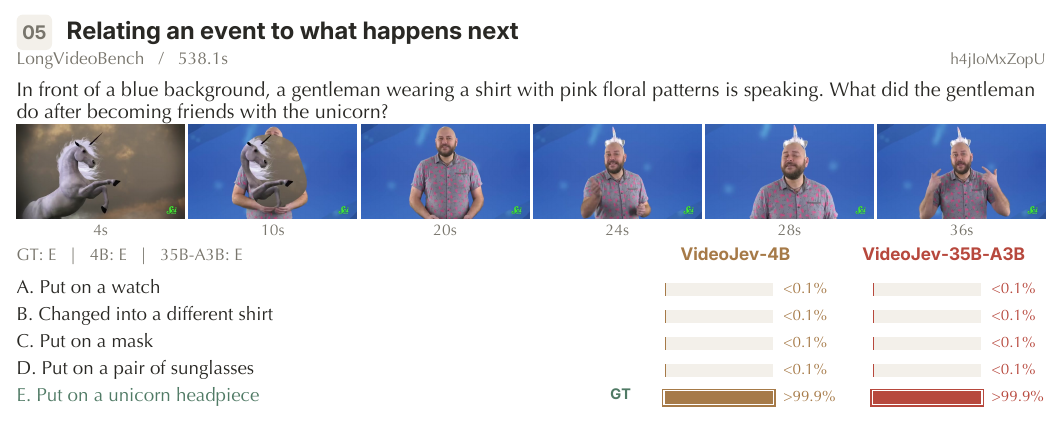}
\caption{Recognizing an event relation on LongVideoBench. Both models select the annotated answer from six options.}
\label{fig:qualitative-05}
\end{figure}

\begin{figure}[!ht]
\centering
\includegraphics[width=\linewidth,trim=0 3bp 0 6bp,clip]{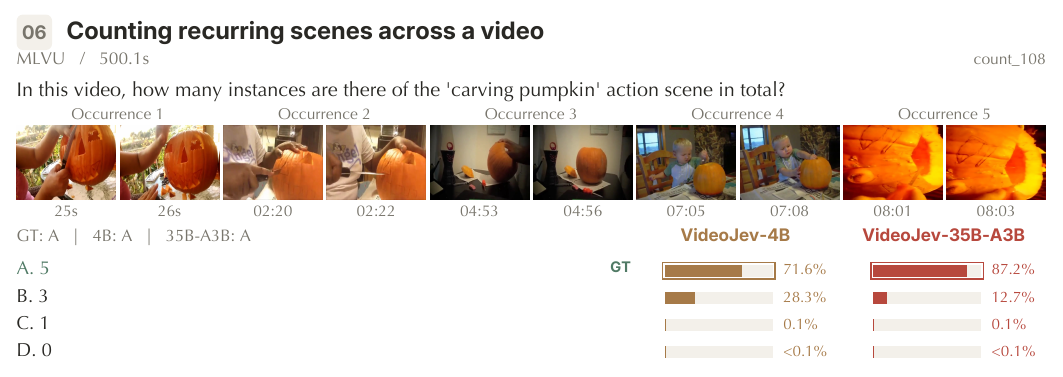}
\caption{Counting five recurring scenes amid unrelated footage on MLVU. Each frame pair shows one occurrence. Both models select the annotated answer, with different option scores.}
\label{fig:qualitative-06}
\end{figure}

\begin{figure}[!ht]
\centering
\includegraphics[width=\linewidth,trim=0 3bp 0 6bp,clip]{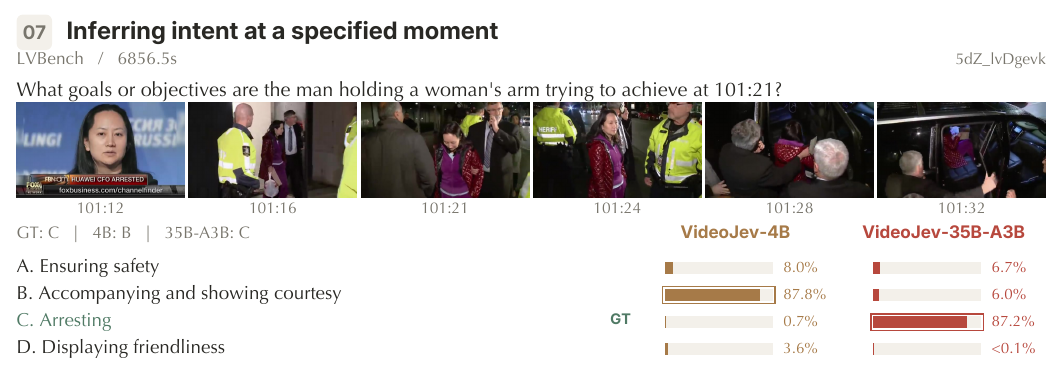}
\caption{An intent-inference disagreement on LVBench: VideoJev-4B interprets the interaction as courtesy, while VideoJev-35B-A3B selects the annotated arrest intent.}
\label{fig:qualitative-07}
\end{figure}

\clearpage
\begin{figure}[!ht]
\centering
\includegraphics[width=\linewidth,trim=0 3bp 0 6bp,clip]{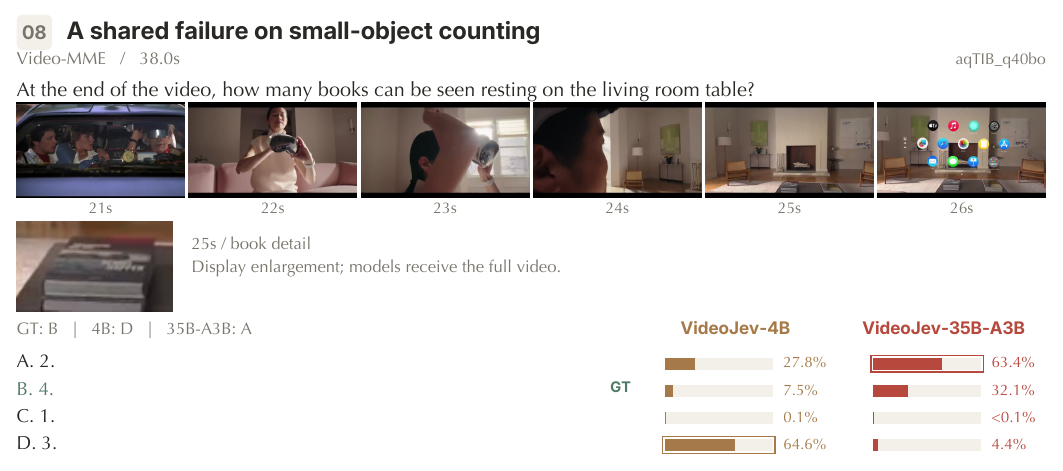}
\caption{A shared failure on small-object counting in Video-MME. Both models miss the annotated answer of four books; the book detail is enlarged for display.}
\label{fig:qualitative-08}
\end{figure}
\endgroup

\end{document}